\documentclass[journal]{IEEEtran}
\usepackage{amsmath,amsfonts}
\usepackage{algorithmic}
\usepackage{array}
\usepackage{textcomp}
\usepackage{stfloats}
\usepackage{url}
\usepackage{verbatim}
\def\BibTeX{{\rm B\kern-.05em{\sc i\kern-.025em b}\kern-.08em
    T\kern-.1667em\lower.7ex\hbox{E}\kern-.125emX}}
\usepackage{balance}

\usepackage{xcolor}
\usepackage{multirow}
\usepackage{booktabs}  
\usepackage{adjustbox} 
\usepackage{makecell}
\usepackage{arydshln}
\usepackage{graphicx}
\usepackage{subcaption}
\graphicspath{{figures/}}
\usepackage{algorithm}

\begin{document}
\title{A Diffusive Data Augmentation Framework for Reconstruction of Complex Network Evolutionary History}
\author{En~Xu,
	Can~Rong,
	Jingtao~Ding,
	and~Yong~Li
	\IEEEcompsocitemizethanks{\IEEEcompsocthanksitem En Xu, Can Rong, Jingtao Ding, and Yong Li are with the Department of Electronic Engineering, Tsinghua University, Beijing, China. \\
		E-mail: xuen@mail.tsinghua.edu.cn, 
		rc20@mails.tsinghua.edu.cn, 
		dingjt15@tsinghua.org.cn,
		liyong07@tsinghua.edu.cn.
	}
	
	\thanks{This work was supported in part by the National Natural Science Foundation of China.\\
		(Corresponding author: Yong Li.)	
}}

\markboth{Journal of \LaTeX\ Class Files,~Vol.~18, No.~9, September~2025}%
{How to Use the IEEEtran \LaTeX \ Templates}

\maketitle

\begin{abstract}
The evolutionary processes of complex systems contain critical information regarding their functional characteristics. The generation time of edges provides insights into the historical evolution of various networked complex systems, such as protein-protein interaction networks, ecosystems, and social networks. Recovering these evolutionary processes holds significant scientific value, including aiding in the interpretation of the evolution of protein-protein interaction networks. 
However, existing methods are capable of predicting the generation times of remaining edges given a partial temporal network but often perform poorly in cross-network prediction tasks. These methods frequently fail in edge generation time recovery tasks for static networks that lack timestamps. In this work, we adopt a comparative paradigm-based framework that fuses multiple networks for training, enabling cross-network learning of the relationship between network structure and edge generation times. Compared to separate training, this approach yields an average accuracy improvement of 16.98\%.
Furthermore, given the difficulty in collecting temporal networks, we propose a novel diffusion-model-based generation method to produce a large number of temporal networks. By combining real temporal networks with generated ones for training, we achieve an additional average accuracy improvement of 5.46\% through joint training.
\end{abstract}

\begin{IEEEkeywords}
Complex network, Evolution, Data Augmentation, Diffusion model.
\end{IEEEkeywords}

\section{Introduction}

\begin{figure*}[t]
	\centering
	\includegraphics[width=0.95\textwidth]{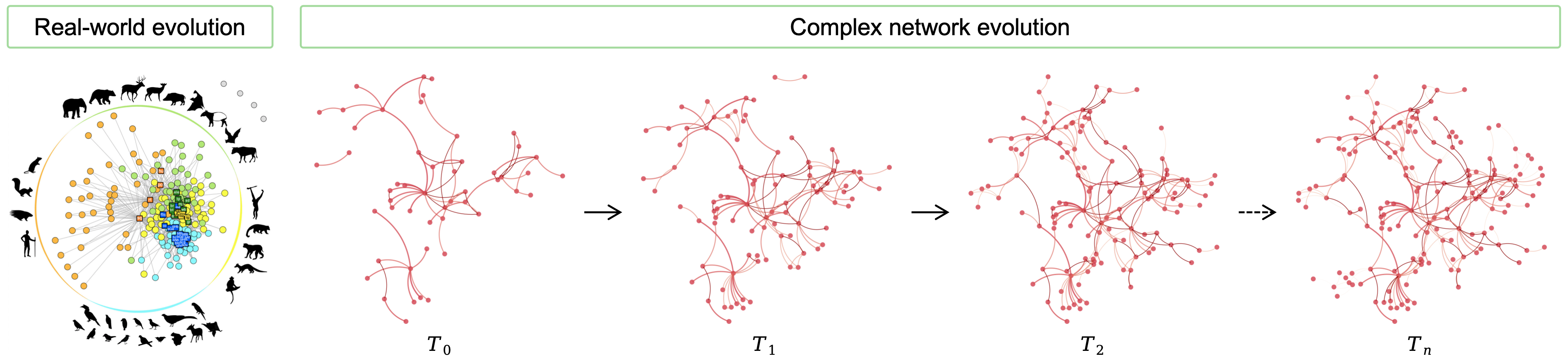}
	\caption{Evolution diagram of a complex network illustrating the dynamic process of node and edge formation over time.}
	\label{fig:evolutionary_schematic}
\end{figure*}

\IEEEPARstart{T}{he} core challenge in reconstructing the evolution of complex networks is inferring the order in which edges are formed in dynamic graphs, thus reproducing the network's evolutionary process (\cite{boccaletti2006complex,liao2017ranking}). This task is intrinsically tied to the dependency between network structure and edge formation sequence, as accurately inferring this order can reveal the underlying dynamic mechanisms of the network, as illustrated in Fig. \ref{fig:evolutionary_schematic}. Successfully reconstructing network evolution has significant applications across various domains, such as social network analysis, traffic system optimization, and biological network research (\cite{barabasi2002evolution,wang2023pattern,seferbekova2023spatial}). These fields require a deep understanding of how networks evolve over time to predict future trends and implement effective interventions or optimizations.

Recent work published in \textit{Nature Communications 2024} \cite{wang2024reconstructing} introduces a novel framework that defines and addresses the important problem of complex network evolution prediction. This approach predicts the generation times of network edges by leveraging the structural features of the network. Specifically, it utilizes the generation times of a subset of edges to infer the temporal information for the remaining edges, effectively linking network structural characteristics with edge generation times. While this represents a significant advancement, many real-world applications involve networks where no temporal information is available, presenting an even more challenging problem.
Many networks are observed as static snapshots without temporal information. For example, in ecological predator-prey networks (as shown in Fig. \ref{fig:evolutionary_schematic}), current interactions are easy to capture, but reconstructing evolutionary trajectories over millennia is costly and challenging. Similarly, in social and protein-protein interaction networks, while current structures are observable, tracing their step-by-step evolution remains difficult.

These scenarios highlight a common challenge in real-world applications: \textbf{inferring the evolutionary history of a network solely from its static structure}. While the \cite{wang2024reconstructing} framework is effective in scenarios where partial temporal information is available, it is not well-suited for tasks where the network is entirely static, with no prior temporal data to leverage. This limitation underscores the necessity of developing novel methodologies capable of reconstructing network evolution directly from static structures, addressing a problem of critical importance across diverse domains.

To address the limitations of existing methods, we reformulate the task of predicting edge generation times into an equivalent task: predicting the relative temporal order between any two edges within the network. This reformulation simplifies the problem and transforms the previously non-transferable task into one that enables cross-network learning of the relationship between network features and edge generation times.

Unlike prior models that are typically trained on a single temporal network, our approach leverages multiple temporal networks for training, allowing the model to learn generalized associations between network features and edge generation times across networks. As a result, the model trained on multiple temporal networks demonstrates superior and consistent performance when applied to previously unseen static networks.

Furthermore, to enhance the model's performance and address the scarcity of real-world temporal network data, we employ a diffusion model to generate an arbitrary number of augmented temporal networks based on existing temporal networks. This augmentation process provides a large and diverse dataset comprising both real and generated temporal networks. By training on this enriched dataset, the model achieves even better performance, particularly in scenarios involving unseen static networks, demonstrating its robustness and effectiveness.

In summary, the key contributions of our work are as follows:

\begin{itemize}
	\item Cross-Network Framework for Edge Generation Time Prediction:
	We introduce a novel framework capable of learning the relationship between network features and edge generation times across multiple temporal networks. This approach enables the model trained on multiple temporal networks to generalize effectively to unseen static networks for edge generation time prediction.
	\item Diffusion-Based Edge Timestamp Generation:
	We propose a diffusion model-based method for generating edge timestamps from network features, achieving exceptional performance in predicting edge generation times and demonstrating robustness and effectiveness.
	\item Integration of Real and Augmented Temporal Networks:
	By integrating real temporal networks with diffusion model-augmented networks, we train a stable, high-performing model that outperforms existing methods in predicting edge generation times in static networks.
\end{itemize}

\section{RELATED WORK} \label{sec:Relatedwork}

\subsection{Temporal Networks in Complex Networks}
Recent advancements in temporal networks have significantly enriched complex network analysis \cite{ding2024artificial,liu2024deep,jin2024large,zhang2024motif,li2024robust}. Early models such as Erdős-Rényi and Barabási-Albert (\cite{karonski1997origins,watts1998collective}) mainly addressed static network properties, offering limited applicability to dynamic systems. \cite{holme2012temporal} provided foundational concepts like temporal paths, establishing a basis for subsequent research. \cite{karsai2012universal} highlighted bursty interaction patterns in real-world networks, emphasizing temporal heterogeneity. \cite{masuda2017random} extended random walk theory to temporal networks, crucial for understanding diffusion processes. \cite{mucha2010community} introduced methods for detecting communities in dynamic structures, while \cite{millidge2024predictive} integrated predictive models for forecasting interactions. \cite{rossetti2018community} reviewed evolving community detection approaches, showcasing the need to address both temporal and topological dynamics in analysis. \cite{wang2024reconstructing} introduced the problem and method for predicting network evolution based on network structures. By utilizing the timestamps of a subset of network edges, the method effectively predicts and reconstructs the remaining edges, achieving promising results. Furthermore, this work demonstrated that the reconstructed edge timestamps could also enhance performance in link prediction tasks, further underscoring the significance of studying network evolution.

\subsection{Dynamic Graphs and Temporal Network Generation}
Recent developments in temporal network generation have significantly advanced the field, focusing on capturing both the structural and temporal dynamics of evolving networks.  \cite {zhou2020data} improved temporal modeling by transforming temporal interactions into static graphs, while \cite{zeno2021dymond} utilized motif-based approaches for generating time-dependent structures, though it assumed fixed rates and lacked adaptability to real-world temporal fluctuations. \cite{gupta2022tigger} introduced TIGGER leveraging temporal point processes, combining both inductive and transductive learning to handle large-scale, complex networks with superior efficiency and accuracy. This progression highlights the growing ability of temporal models to replicate the nuanced evolution of real-world dynamic systems.

\subsection{Data Augmentation via Graph Diffusion Models}
Employing generative methods such as generative adversarial networks~(GANs) to learn a distribution based on a small sample of data and subsequently conduct sampling to obtain a large quantity of training samples is a commonly adopted method~\cite{zhang2022multi,abdollahzadeh2023survey,zhang2018metagan,fu2024dreamda}. Diffusion models have recently shown significant advantages in performance, and have done well on graph generation tasks~\cite{vignac2022digress,niu2020permutation,jo2022score,liu2024tdnetgen}, making them suitable for solving the problem of small-sample learning in reconstructing network evolutions. Specifically, \cite{niu2020permutation} first propose to introduce the score-based diffusion model to solve the graph generation problem. The Gaussian noise is adopted to construct the diffusion process, leading to poor sparsity generated graphs. \cite{jo2022score} explore jointly modeling the nodes and edges with semantic features. \cite{vignac2022digress} use multinomial noise to make the diffusion process operating in discrete space, which greatly improves the performance. Therefore, we can see that using graph diffusion models to generate sufficient samples to enhance edge order prediction performance is promising.
\section{Preliminaries} \label{sec:pre}
\subsection{Reformulating the Edge Generation Prediction Task}
In the study of network evolution, predicting the exact generation time of edges represents a significant challenge. Directly learning the relationship between network features and edge generation times proves to be difficult due to the inherent complexity and variability of these relationships across different networks. Moreover, such a task lacks generalizability; the features learned in one network often fail to generalize effectively to another, rendering the task unsuitable for cross-network transferability. Thus, learning to predict edge generation times within the same network—where partial timestamps are available for predicting the remaining ones—is appropriate. However, if the goal is to generalize across networks and learn the relationship between network features and edge generation times, the equivalence transformation becomes crucial.

To address this limitation, we reformulate the original task of predicting edge generation times into an equivalent task: predicting the relative temporal order of generation between any two edges in the network. This reformulation simplifies the prediction problem into a pairwise classification task, which is more robust and transferable across networks.

We rigorously prove the mathematical equivalence between these two tasks, with the final equivalence result expressed as:
\begin{equation}
	\mathcal{E}_\text{theory} = \sqrt{\frac{x(1-x)}{(2x-1)^2}} \frac{1}{\sqrt{M}}, \label{eq:equivalence}
\end{equation}
where \(x\) represents the pairwise prediction accuracy, and \(M\) denotes the total number of edges in the network. The detailed proof is provided in Appendix \ref{apdx:equal}. 

\begin{figure}[t]
	\centering
	\includegraphics[width=0.4\textwidth]{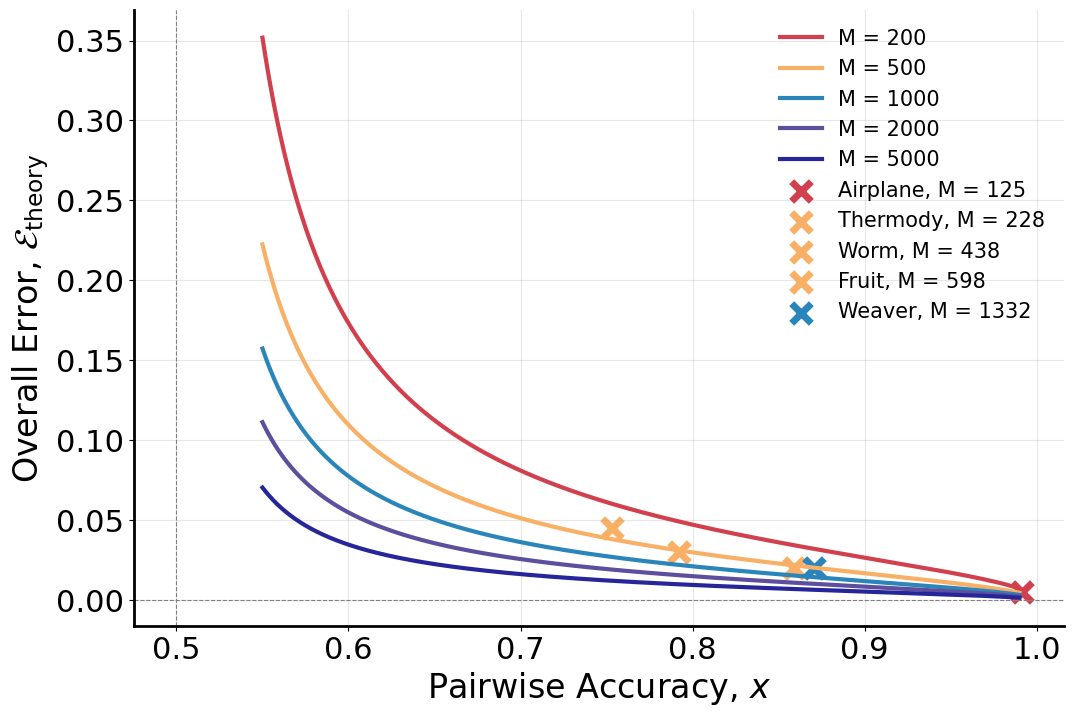}
	\caption{Equivalence transformation in network evolution time prediction tasks.}
	\label{fig:rmse_acc}
\end{figure}

Furthermore, as illustrated in Fig. \ref{fig:rmse_acc}, experimental results validate this equivalence in real-world networks. These results demonstrate that improving pairwise prediction accuracy directly contributes to reducing the overall prediction error in edge generation times. Equation \ref{eq:equivalence} reveals that the overall sequence error scales inversely with the square root of the edge count, favoring networks with many edges. 

This equivalence also yields an additional benefit: achieving a low RMSE in predicting edge generation times does not necessitate exceedingly high accuracy in pairwise temporal order predictions. This insight reduces the performance requirements for the underlying pairwise prediction model, enhancing the practical applicability of the approach.

To infer the complete temporal order of edge generation from pairwise relationships, we adopt the Borda algorithm. This algorithm aggregates pairwise predictions to construct a global sequence of edge generation times while ensuring consistency across predictions. A detailed explanation of the Borda algorithm, including its derivation and implementation, is provided in the Appendix \ref{apdx:borda}. 

\subsection{Mathematical Definition of the Task}
In the study of complex networks, the generation time of edges is a critical factor influencing network evolution. Our prediction task aims to learn the relationship between network structure and edge generation time, given a temporal network. Specifically, we focus on how to infer the generation order of edges in the absence of time labels.

Our ultimate goal is to predict the generation time of each edge for a pure network structure. To simplify the task, we specify it as follows: for any two edges \( e_i \) and \( e_j \) in the network, we need to predict their generation order, determining which edge is generated first and which is generated later. By inferring the order of all edges, we can construct the overall generation sequence of edges in the network, thereby gaining a comprehensive understanding of the network's evolutionary process.

1. \textit{Temporal Network Representation}:
Let the temporal network \( G \) consist of a set of nodes \( V \) and a set of edges \( E \). Each edge \( e_k \in E \) represents a connection between a pair of nodes \( (u_k, v_k) \) and has an associated generation time \( t_k \).

2. \textit{Edge Generation Order}:
For any two edges \( e_i = (u_i, v_i) \) and \( e_j = (u_j, v_j) \), we define the generation order relation \( R(e_i, e_j) \) as follows:
\[
R(e_i, e_j) = 
\begin{cases} 
	1, & \text{if } t_i < t_j \\ 
	0, & \text{if } t_i \geq t_j 
\end{cases}
\]
Here, \( R(e_i, e_j) = 1 \) indicates that edge \( e_i \) is generated before edge \( e_j \), while \( R(e_i, e_j) = 0 \) indicates that the generation time of edge \( e_i \) is later than or equal to that of edge \( e_j \).

3. \textit{Edge Generation Time Prediction}:
Our task can be expressed as a binary relation prediction problem, where the objective is to predict the edge generation order \( R(e_i, e_j) \) by learning the structural features of the network \( S(G) \):
\[
\hat{R}(e_i, e_j) = f(S(G), e_i, e_j)
\]
In this context, \( \hat{R}(e_i, e_j) \) is the predicted generation order, and the function \( f \) represents the model used to infer the generation order based on network structural features.

4. \textit{Construction of Edge Generation Sequence}:
Once the generation order for all edges is predicted, we can obtain the overall edge generation sequence \( T \), expressed as:
\[
T = \{ e_{k_1}, e_{k_2}, \ldots, e_{k_m} \}
\]
Here, \( k_1, k_2, \ldots, k_m \) are the indices of edges arranged according to the predicted generation order.

Through the above description and mathematical definitions, our task clearly demonstrates how to infer edge generation times based solely on structural information in the absence of time labels, thus providing robust support for the evolutionary prediction of complex networks.

\subsection{Framework for Comparative Paradigm-based Neural Network}
\begin{figure*}[t]
	\centering
	\includegraphics[width=0.8\textwidth]{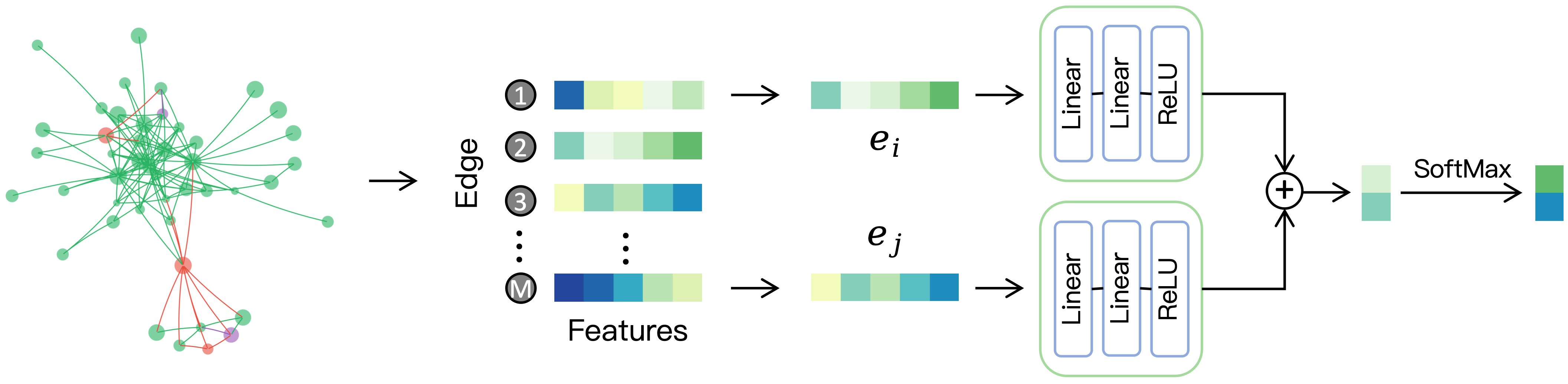}
	\caption{Schematic framework of comparative paradigm-based neural network (CPNN).}
	\label{fig:CPNN}
\end{figure*}
The framework for predicting the order of two edges in a network is structured as follows. Initially, we utilize an embedding representation learning method to obtain a representation vector for each edge in the network. These embedding vectors are derived from 12 network features, which include the degree, clustering coefficient, betweenness centrality, common neighbors, random walk probability, minimum spanning tree connection, Jaccard coefficient, resource allocation index, Adamic-Adar index, shortest path length, PageRank, and average clustering coefficient \cite{lu2011link,hesamipour2019new,bianchini2005inside}. These features are calculated based on the network’s adjacency matrix and provide a comprehensive structural description of each edge. Let us denote the edges as \( e_1 \) and \( e_2 \), with their respective embedding vectors represented as \( \mathbf{h}_1 \) and \( \mathbf{h}_2 \).

Each embedding vector is subsequently fed into a fully connected neural network comprising three layers. The architecture is defined as follows:

Input Layer: The first layer receives the input vectors \( \mathbf{h}_1 \) and \( \mathbf{h}_2 \), which are concatenated to form a single input vector \( \mathbf{h} = [\mathbf{h}_1; \mathbf{h}_2] \in \mathbb{R}^{d} \), where \( d \) is the dimension of the embedding vectors.

Hidden Layer: The second layer is a hidden layer with a dimensionality of \( \frac{2}{3}d \). This layer applies a linear transformation followed by a ReLU activation function, defined as:
\[
\mathbf{h}_{\text{hidden}} = \text{ReLU}(\mathbf{W}_1 \mathbf{h} + \mathbf{b}_1),
\]
where \( \mathbf{W}_1 \in \mathbb{R}^{\frac{2}{3}d \times d} \) and \( \mathbf{b}_1 \in \mathbb{R}^{\frac{2}{3}d} \) are the weight matrix and bias vector for the hidden layer, respectively.

Output Layer: The output layer consists of two neurons that produce a scalar output \( z_1 \) and \( z_2 \):
\[
\begin{bmatrix}
	z_1 \\
	z_2
\end{bmatrix} = \mathbf{W}_2 \mathbf{h}_{\text{hidden}} + \mathbf{b}_2,
\]
where \( \mathbf{W}_2 \in \mathbb{R}^{2 \times \frac{2}{3}d} \) and \( \mathbf{b}_2 \in \mathbb{R}^{2} \) are the weights and biases for the output layer.

After obtaining the outputs \( z_1 \) and \( z_2 \), we apply the softmax function to convert these into a probability distribution representing the likelihood that each edge was generated first:
\[
p(e_1 \text{ before } e_2) = \frac{e^{z_1}}{e^{z_1} + e^{z_2}},
\]
\[
p(e_2 \text{ before } e_1) = \frac{e^{z_2}}{e^{z_1} + e^{z_2}}.
\]

These probabilities reflect the generation times of the edges, normalized to ensure that they sum to one.

The loss function \( L \) is defined as the categorical cross-entropy loss between the predicted probabilities and the true labels \( y \in \{0, 1\} \), where \( y = 1 \) indicates that \( e_1 \) occurs before \( e_2 \):
\[
L = -y \log(p(e_1 \text{ before } e_2)) - (1 - y) \log(p(e_2 \text{ before } e_1).
\]

Additionally, we incorporate $L2$ regularization to mitigate overfitting. The regularization term is defined as:
\[
R = \lambda \left( \|\mathbf{W}_1\|^2 + \|\mathbf{W}_2\|^2 \right),
\]
where \( \lambda \) is the regularization strength. Thus, the final loss function to be minimized is:
\[
\mathcal{L} = L + R.
\]

\begin{figure}[t]
	\centering
	\includegraphics[width=0.25\textwidth]{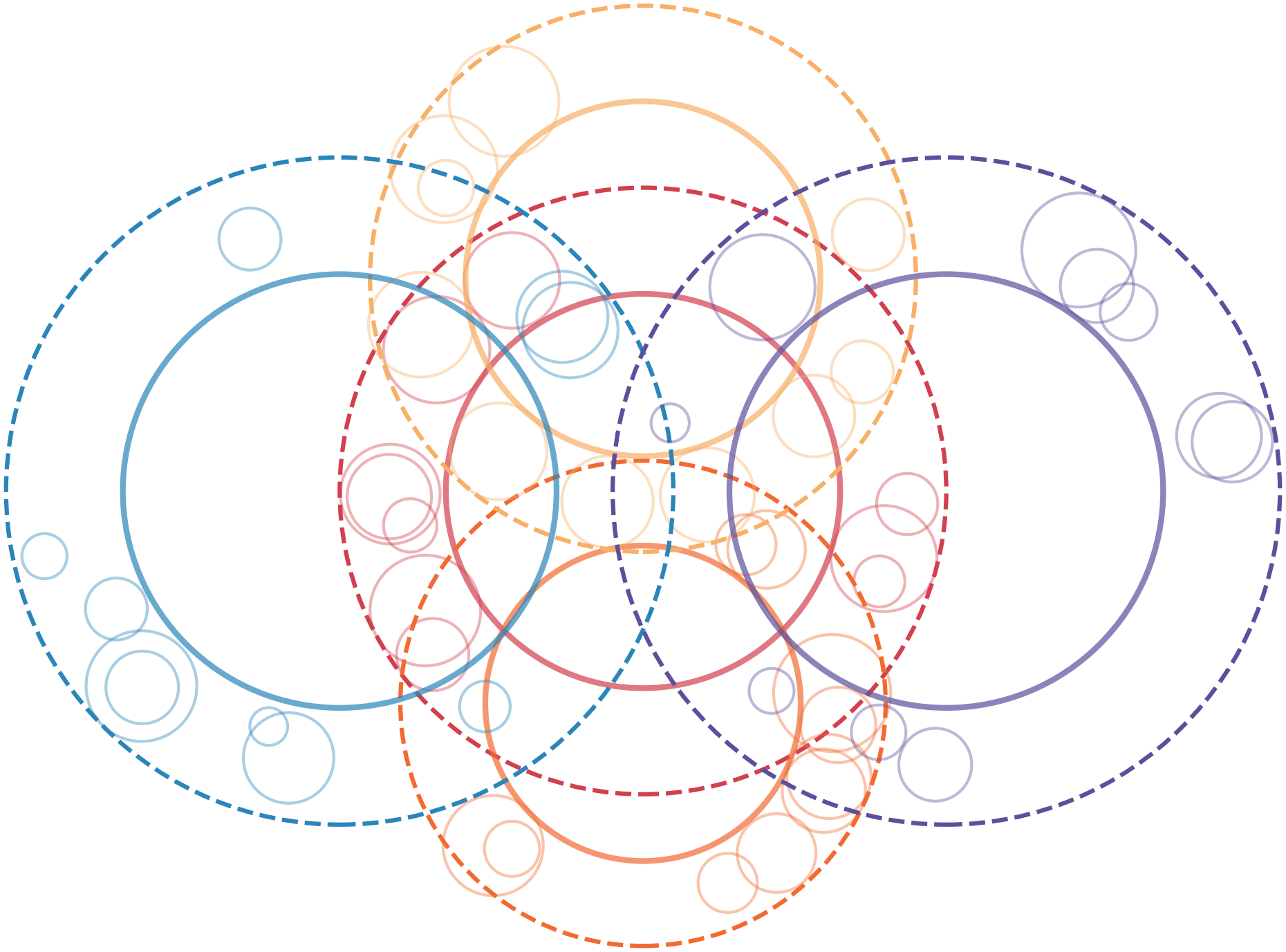}
	\caption{Illustration showing the challenges of traditional methods in transfer learning, how combined training improves transfer performance, and how sample augmentation based on diffusion models enhances the model's transferability and accuracy.}
	\label{fig:enhance_diagram}
\end{figure}
\subsection{Limitations of Existing Method} 
Current methods for network evolution prediction rely on partial edge timestamps to train models and predict the temporal order of remaining edges. For instance, the Nature Communications \cite{wang2024reconstructing} approach performs well on synthetic networks by learning from one fully timestamped network and predicting on another. However, its effectiveness diminishes on real-world networks due to the structural homogeneity of synthetic networks (e.g., Barabási–Albert \cite{barabasi1999mean}, Popularity-similarity-optimization model \cite{papadopoulos2014network}, Fitness model \cite{bianconi2001competition}), which share scale-free properties. In contrast, real-world networks exhibit diverse and complex generative mechanisms, making learned models difficult to generalize across networks.

A more critical limitation arises in scenarios where no temporal information is available. In practice, collecting edge timestamps is often infeasible, leaving only the network's topological structure. Existing methods, which depend on partial temporal annotations, fail to address the challenge of predicting temporal dynamics in completely untimestamped networks.

\section{Method} \label{sec:methods}

\begin{figure*}[t]
	\centering
	\includegraphics[width=\textwidth]{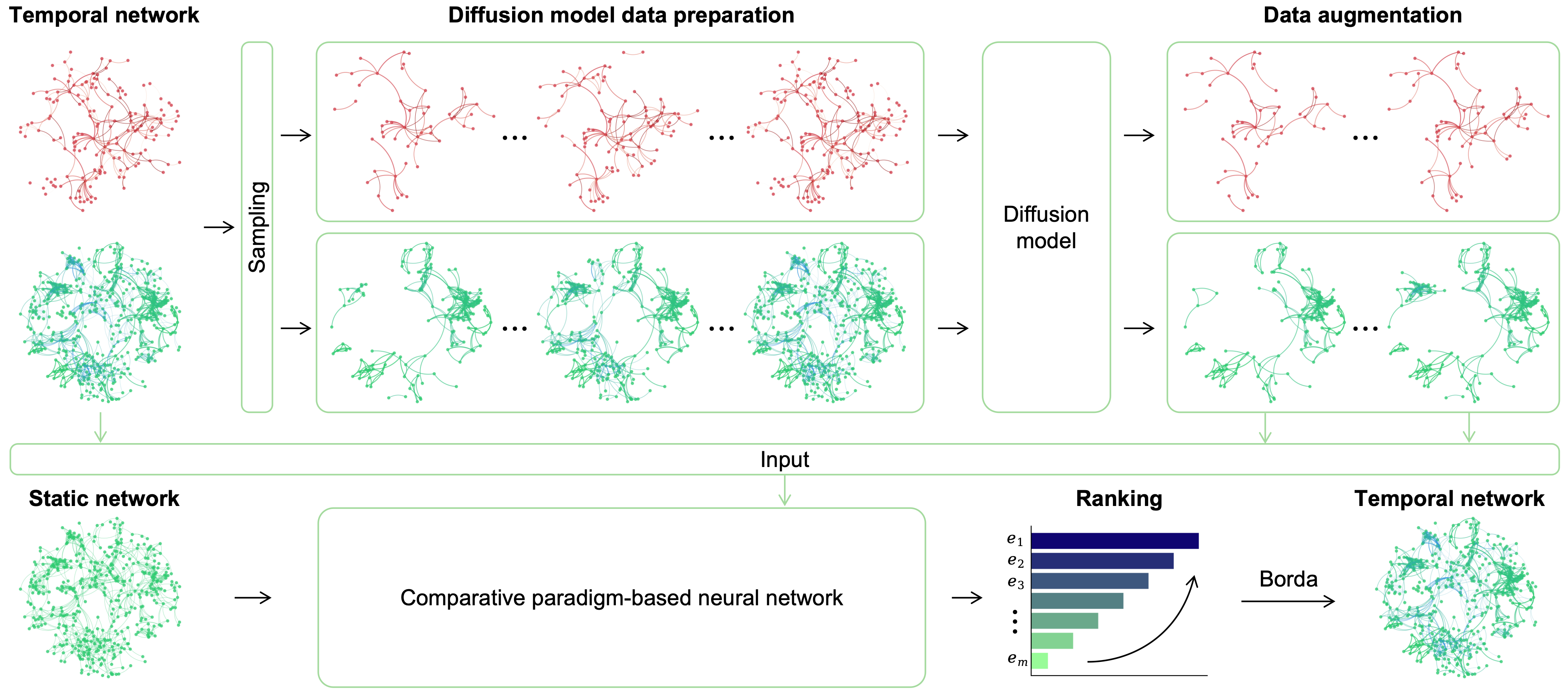}
	\caption{Illustration of the framework for predicting the edge evolution process of static networks based on diffusion model-driven data augmentation.}
	\label{fig:framework}
\end{figure*}
\subsection{Joint Training for Improved Transferability}
To address the limitations of existing methods, we propose a joint training approach that combines multiple networks to train a single model, exposing it to diverse network types and improving its ability to generalize to unseen networks.

In Section \ref{exp:benefit}, we demonstrate why traditional methods struggle with transferability when trained on a single network type. Structural differences across networks lead to significant performance gaps, making these models unreliable on unseen data. Reformulating the prediction task—from predicting edge generation times to predicting pairwise edge order—mitigates this issue. Furthermore, Section \ref{exp:multi} highlights the superior performance of our jointly trained model, which achieves consistent results on unseen networks.

Fig.~\ref{fig:enhance_diagram} illustrates our approach. Solid circles represent existing networks, and current methods often exhibit instability, as gaps between networks hinder generalization. Joint training reduces these gaps by incorporating a diverse range of networks, enabling the model to learn more generalizable patterns.

To further enhance the model’s transferability, we generate augmented networks using diffusion models, represented by the small solid circles in Figure~\ref{fig:enhance_diagram}. These augmented networks expand the range of each original network, increasing coverage across network types and reducing gaps. The broader state space allows the model to encounter more diverse network structures during training, improving its robustness on unseen networks. The dashed circles in Figure~\ref{fig:enhance_diagram} represent the combined sets of original and augmented networks, showcasing the scalability and adaptability of our approach.

In Section \ref{method:diffusion}, we detail the diffusion-based augmentation process and demonstrate its effectiveness in further improving the model’s performance.

\subsection{Enhancing Network Evolution Prediction with Diffusion Model-Based Augmentation}
As shown in Fig. \ref{fig:framework}, we propose an enhanced framework for network evolution prediction based on diffusion models. The core idea is that we have a limited number of temporal networks. By performing extensive sampling on each temporal network, we input the samples into the diffusion model to learn the underlying generation mechanism of the temporal networks. The diffusion model then generates an arbitrary number of augmented samples. Using a combination of a small number of real temporal networks and a large number of diffusion-augmented temporal networks, we input them into a comparative paradigm-based neural network (CPNN) for training. This approach results in a model with better transferability and stability. In practical applications, when a static network is input into the trained CPNN, we can predict the temporal order of pairs of edges. Further, by applying the Borda count algorithm, we can derive the overall edge order of the entire network, enabling the prediction of edge evolution processes in static networks.

\subsubsection{Construction of Training Dataset for the Diffusion Model}\label{method:diffusion}

The construction of the training dataset for our diffusion model is a crucial step in enhancing the edge order prediction task. Our approach begins with a labeled temporal network, which serves as the foundational dataset. The methodology can be outlined as follows:

Sampling Strategy: We employ a systematic sampling technique to generate multiple training network samples from the original labeled temporal network. Specifically, we start by sampling a subset of edges, beginning with 50\% of the total edges and progressively increasing to include all edges in the network. This approach allows us to explore various network densities and their impacts on the learning process.

Creation of Diverse Samples: For each sampling iteration, we construct a new training network by retaining the sampled edges and discarding the remaining edges. This results in 100 distinct training network samples, each representing a unique configuration of the original temporal network. As shown in Fig. \ref{fig:framework}, we sampled $G^1, \dots, G^{100}$ networks. By varying the percentage of edges retained, we ensure a diverse representation of network structures, which is essential for robust model training.

\begin{figure*}[t]
	\centering
	\includegraphics[width=0.9\textwidth]{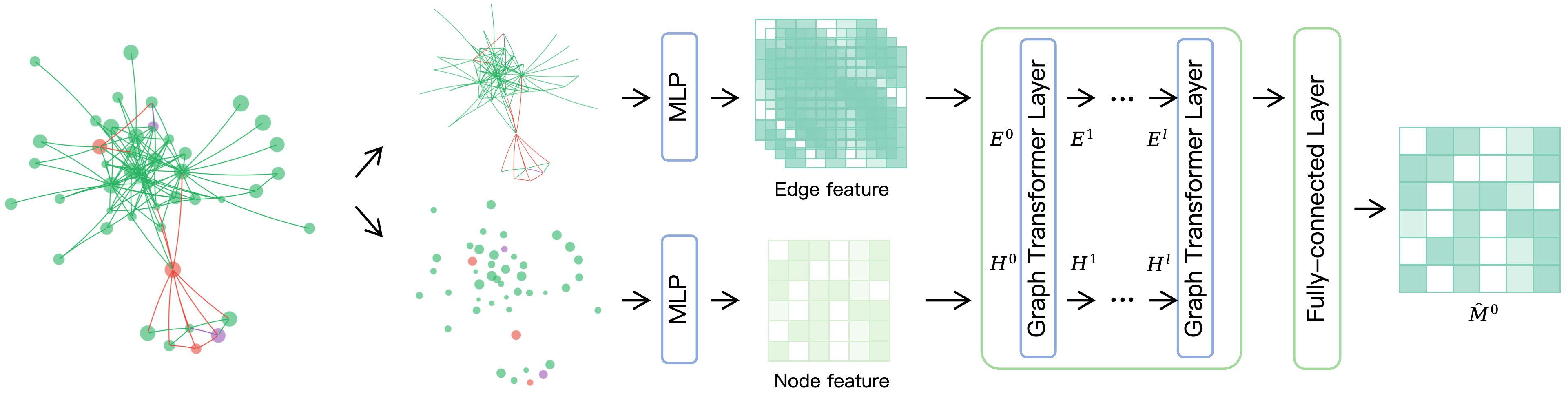}
	\caption{Framework diagram of the predictor in the denoising process of TopoEvoDiff.}
	\label{fig:TopoEvoDiff}
\end{figure*}
\subsubsection{Graph Denoising Diffusion Model}
Based on the training data sampled from the real temporal network, we utilize the graph diffusion models to learn the distribution of the topology evolutions conditioned on the final topology structure. We refer to our model as the \textit{Diffusion Model for Generating Topology Evolution History}, abbreviated as \textbf{TopoEvoDiff}. The graph diffusion models are adapted to ordered edges diffusion based on the final topology structure, where the structural features are encoded into the node features, such as \textcolor{black}{node embeddings learned through structural learning}.

The graph diffusion model consists of two main procedures: the forward diffusion process $q$ and the reverse denoising process $p$. In the forward diffusion process, the network orders are disrupted by Gaussian noise step by step and reach to pure noise. The forward diffusion process is utilized to construct the training data for the denoising networks used in the reverse denoising process. The computation is defined as: \textcolor{black}{$F$ should mean the timestamps and $ij$ should mean the indices of the edges}
\begin{equation}\label{eq:ODdiff}
    \begin{split}
    \begin{aligned}
    & q(F_{ij}^t|F_{ij}^{t-1})=\mathcal{N}(F_{ij}^t; \sqrt{1-\beta_t} F_{ij}^{t-1} , \beta_t \mathbf{I}),\\
    & q(F_{ij}^1, ..., F_{ij}^T|F_{ij}^0) = \prod_{t=1}^T {q(F_{ij}^t|q^{t-1})}.
    \end{aligned}
    \end{split}
\end{equation}
where $t$ represents the diffusion steps, $\mathcal{N}$ denotes the normal distribution, $\beta_t$ indicates the noise level at step \(t\), and \(\mathbf{I}\) refers to the identity matrix.
The reverse denoising process is to recover the origin edge orders from the pure noise, with the guidance of the final topology structure. In each step, the noised to be removed from the noisy data is predicted by the denoising networks. The denoising networks used in the reverse process are graph transformers~\cite{dwivedi2020generalization}. The construction of the prediction model during the denoising process is shown in Fig. \ref{fig:TopoEvoDiff}. In the denoising process, node features serve as guidance to direct the data generation, remaining unchanged throughout. In each step of the denoising process, the noisy edges and the node features used as guidance are fed into the graph transformer as the node inputs and edge inputs, and the small noise that needs to be removed from the noisy edges in a single step is output. The computation of the reverse denoising process is defined as:
\begin{equation}
    \begin{split}
    \begin{aligned}
    & p_{\theta} (\mathbf{F}^{t-1}|\mathbf{F}^t,\mathcal{C_\mathcal{R}}) = \mathcal{N} (\mathbf{F}^{t-1} ; \mu_{\theta}(\mathbf{F}^t,t,\mathcal{C_\mathcal{R}}) ,(1-\bar{\alpha}^t)\mathbf{I}),\\
    \end{aligned}
    \end{split}
\end{equation}
where
\begin{equation}
    \begin{split}
    \begin{aligned}
    & \mu_{\theta}(\mathbf{F}^t,t,\mathcal{C_\mathcal{R}}) = \frac{1}{ \sqrt{\alpha_t} } (\mathbf{F}^t - \frac{\beta_t}{\sqrt{1-\bar{\alpha}_t}} \mathbf{\epsilon}_\theta(\mathbf{F}^t,t,\mathcal{C_\mathcal{R}}) ) ,\\
    \end{aligned}
    \end{split}
\end{equation}
$ \alpha_t = 1-\beta_t$, and $\bar{\alpha}_t = \prod_{i=1}^t \alpha_i$. Here, $\mathbf{\epsilon}_\theta(\mathbf{F}^t,t,\mathcal{C_\mathcal{R}})$ is the noise predicted by $\theta$ based on the noisy state $\mathbf{F}^t$, diffusion step $t$ and the spatial characteristics $\mathcal{C_\mathcal{R}}$ of the topology of the node

The denoising networks are trained to minimize the reconstruction loss between the predicted noise and the true noise. The loss function we adopted is the same as denoising diffusion probabilistic models~(DDPMs)~\cite{ho2020denoising}, which is the mean squared error between the predicted noise and the true noise. The loss function is defined as:
\begin{equation}
    \begin{split}
    \begin{aligned}
        \mathcal{L} = \mathbb{E}_{t,\epsilon \sim \mathcal{N}(0,\mathbf{I})} \left[ \| \mathbf{\epsilon} - \mathbf{\epsilon}_\theta(\mathbf{F}^t,t,\mathcal{C_\mathcal{R}}) \|_2^2 \right]
    \end{aligned}
    \end{split}
\end{equation}
where $\| \dot \|$ denotes the $L2$ norm. 

When the model is trained, we can sample the temporal networks with similar distribution as the original dataset by the graph diffusion model. First, a pure Gaussian noise is sampled and then the denoising networks iteratively predict the noise to be removed, and the ordered edges will be obtained from the weights of the sampled network gradually. The generated temporal networks are used to augment the original dataset for the edge order prediction model.
\section{Experiments} \label{sec:exp}
\subsection{Datasets}
This study employs 10 real-world datasets from five distinct categories: Protein-Protein Interaction (PPI) Networks, World Trade Web (WTW), Collaboration Networks, Animal Networks, and Transportation Networks. These datasets, covering diverse domains, provide a solid foundation for evaluating the proposed methods. The details of the datasets are summarized in Table~\ref{tab:dataset-info}, and the descriptions are as follows:
\begin{table*}[htbp]
	\centering
	\caption{Summary of the 10 real-world networks. Columns include: network category, network name, number of nodes (\(N\)), number of edges in the final snapshot (\(M\)), number of edge pairs that distinguish the generation order (\(E_d\)), the proportion of distinguishable edge pairs (\(P_{E_d} = E_d / \binom{M}{2}\)), number of snapshots (\(S\)), and the reference for each dataset.}
	\label{tab:dataset-info}
	\scalebox{1}{
		\begin{tabular}{llrrrrrr}
			\toprule
			\textbf{Network type} & \textbf{Network name} & \textbf{N} & \textbf{M} & \textbf{$E_d$} & \textbf{$P_{E_d}$} & \textbf{S} & Refs \\
			\hline
			\multirow{5}{*}{Protein-Protein Interaction (PPI)} & Fungi & 2,144 & 6,000 & 3,132,599 & 0.174 & 3 & \multirow{5}{*}{\cite{jin2013evolutionary}}\\
			& Human & 1,891 & 2,840 & 1,467,231 & 0.364 & 3 &\\
			& Fruit & 461 & 598 & 58,768 & 0.329 & 3 &\\
			& Worm & 485 & 438 & 57,980 & 0.606 & 4 &\\
			& Bacteria & 873 & 2,321 & 357,730 & 0.133 & 2 &\\
			\hline
			\multirow{1}{*}{World Trade Web} & WTW & 187 & 3,249 & 1,799,895 & 0.341 & 17 &\multirow{1}{*}{\cite{garcia2016hidden}}\\
			\hline
			\multirow{1}{*}{Collaboration} & Thermody & 158 & 228 & 25,668 & 0.992 & 131 &\multirow{1}{*}{\cite{hu2014conditions}}\\
			\hline
			\multirow{1}{*}{Animal} & Weaver & 445 & 1,332 & 762,058 & 0.860 & 8 &\multirow{1}{*}{\cite{rossi2015network}}\\
			\hline
			\multirow{2}{*}{Transportation} & Airplane & 48 & 125 & 1,291 & 0.167 & 5 &\multirow{2}{*}{\cite{gallotti2015multilayer}}\\
			& Coach & 1,874 & 2,666 & 84,495 & 0.024 & 4 &\\
			\bottomrule
		\end{tabular}
	}
\end{table*}

\begin{itemize}
	\item PPI Networks: This category includes five networks: Fungi, Human, Fruit, Worm, and Bacteria. Each node represents a protein, while edges denote interactions between proteins. The datasets capture the first occurrence of each edge to distinguish the order in which interactions are established. These networks are particularly valuable for studying temporal evolution in biological systems.
	
	\item WTW: The WTW dataset represents global trade relations between countries from 1997 to 2013. Nodes correspond to countries, and edges signify bilateral trade relationships. The generation time of an edge reflects when trade between two countries was first recorded. This dataset highlights the evolving dynamics of global trade.
	
	\item Collaboration Networks: We use the Thermody collaboration network, constructed from publication data of the American Physical Society. Nodes represent researchers, and edges denote coauthorship relationships. The earliest collaboration between two authors determines the generation time of an edge. This dataset provides insight into academic collaboration patterns.
	
	\item Animal Networks: The Weaver network is based on 10 months of social interaction data from colonies of weaver birds in South Africa. Nodes represent individual birds, and edges denote interactions. To improve data density and reduce sparsity, the raw data is aggregated into 8 snapshots, capturing the temporal dynamics of social behavior.
	
	\item Transportation Networks: Two transportation networks are used: Airplane and Coach. In the Airplane network, nodes are airports, and edges represent flight connections, while in the Coach network, nodes are coach stations, and edges denote transport links. The generation time of an edge is the earliest recorded connection between two nodes. The data is aggregated into daily snapshots, resulting in 5 snapshots for Airplane and 4 snapshots for Coach.
\end{itemize}

These datasets, spanning diverse domains such as biology, economics, collaboration, animal, and transportation, provide comprehensive support for the experiments in this study.

\subsection{Baseline Models}

To evaluate the effectiveness of our proposed method, we compare it against a set of state-of-the-art baseline models. These models are designed to explore the relationship between node, edge, and network features and the edge weights within a network. The baseline models fall into three categories and are described as follows:

\begin{itemize}
		\item Random Forest (RF) \cite{pourebrahim2019trip}: A tree-based machine learning algorithm that uses node and edge-level features to predict edge weights. Its robustness and ability to handle non-linear feature interactions make it a strong benchmark.
		\item TIGGER \cite{gupta2022tigger}: A temporal generative model that combines synthetic network generation with node embeddings to capture dynamic relationships and predict edge weights over time.
		\item NetGAN \cite{bojchevski2018netgan}: A generative adversarial network (GAN)-based model designed to recreate realistic network structures by learning the patterns of walks in the graph. The model is adapted here to generate networks with weighted edges.
		\item Digress \cite{vignac2023digress}: A state-of-the-art method for network generation, utilizing a discrete denoising diffusion model to directly operate on graph structures while preserving their sparsity and topology. By employing a Markovian noise process and a graph transformer for reconstruction, Digress sets a new benchmark in generating large-scale graphs.
\end{itemize}

These baselines encompass a wide range of methodologies, from classical physics-based models to advanced deep learning and generative approaches, providing a comprehensive basis for comparison. The detailed parameter settings for these models are presented in the supplementary material.

\begin{figure*}[htbp]
	\centering
	\caption{Performance on Training Networks with Augmentation.}
	\label{fig:augmentation_self}
	\begin{subfigure}{0.3\textwidth}
		\centering
		\includegraphics[width=\linewidth]{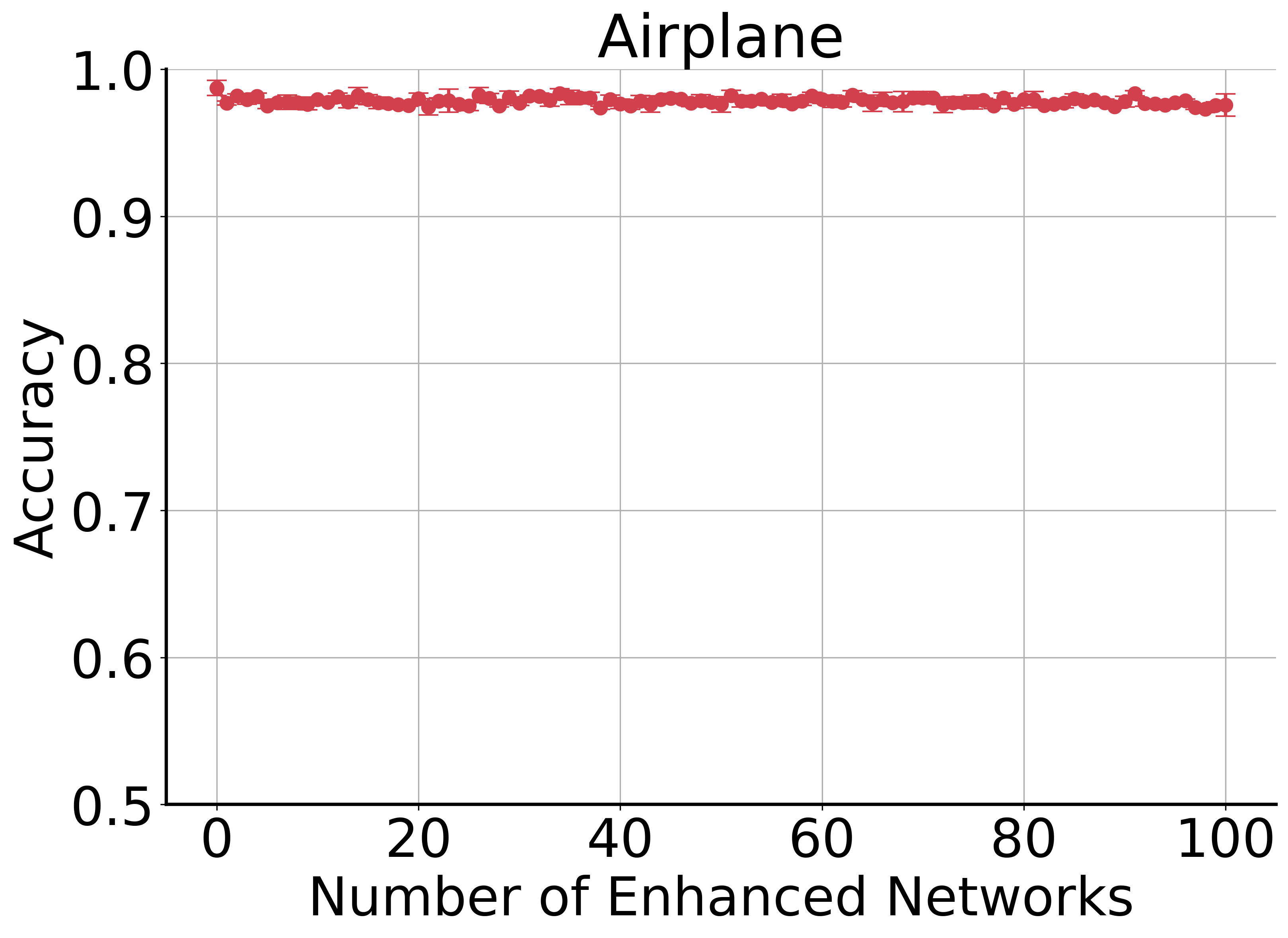}
		\caption*{} 
	\end{subfigure}
	\begin{subfigure}{0.3\textwidth}
		\centering
		\includegraphics[width=\linewidth]{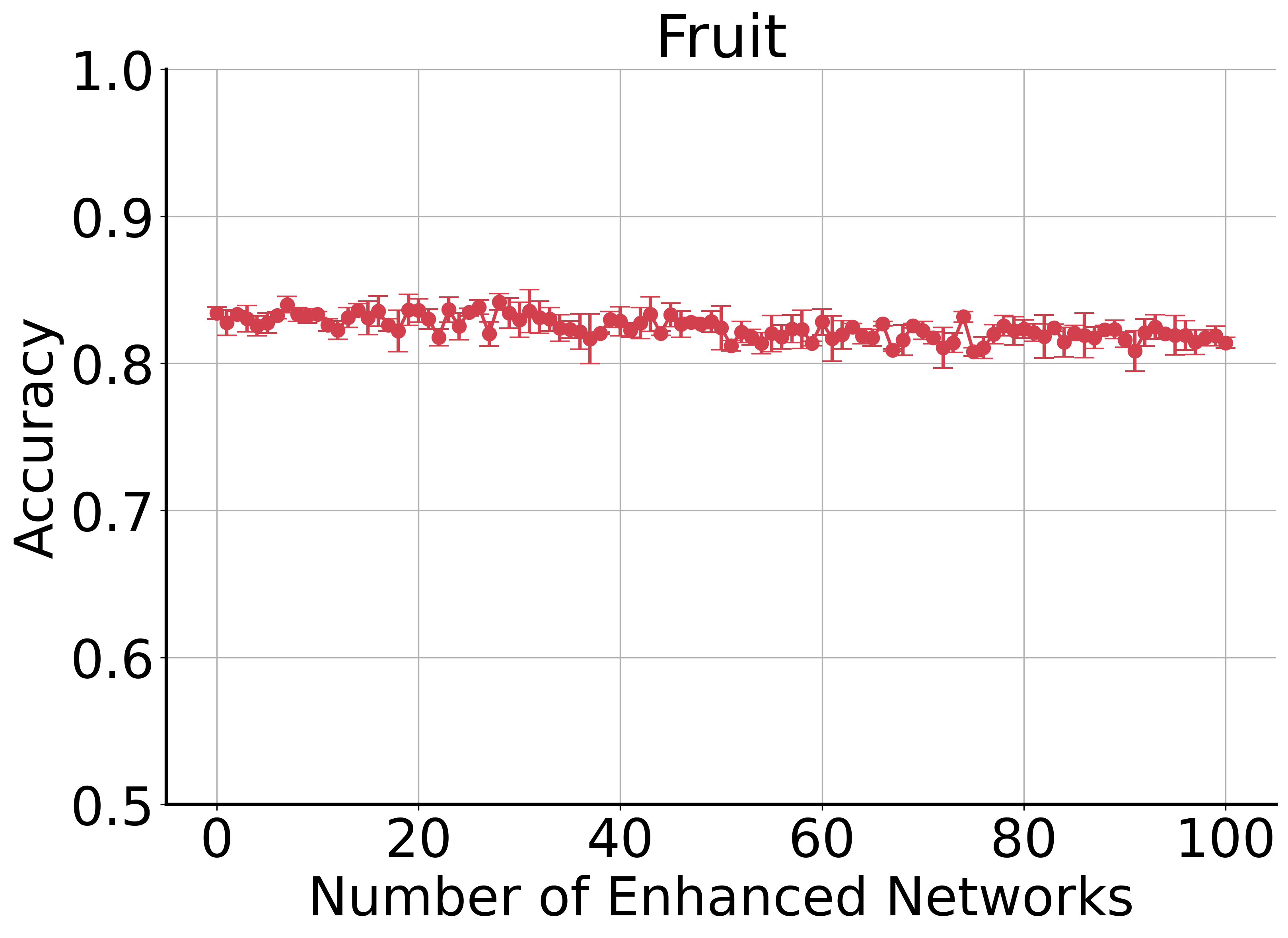}
		\caption*{} 
	\end{subfigure}
	\begin{subfigure}{0.3\textwidth}
		\centering
		\includegraphics[width=\linewidth]{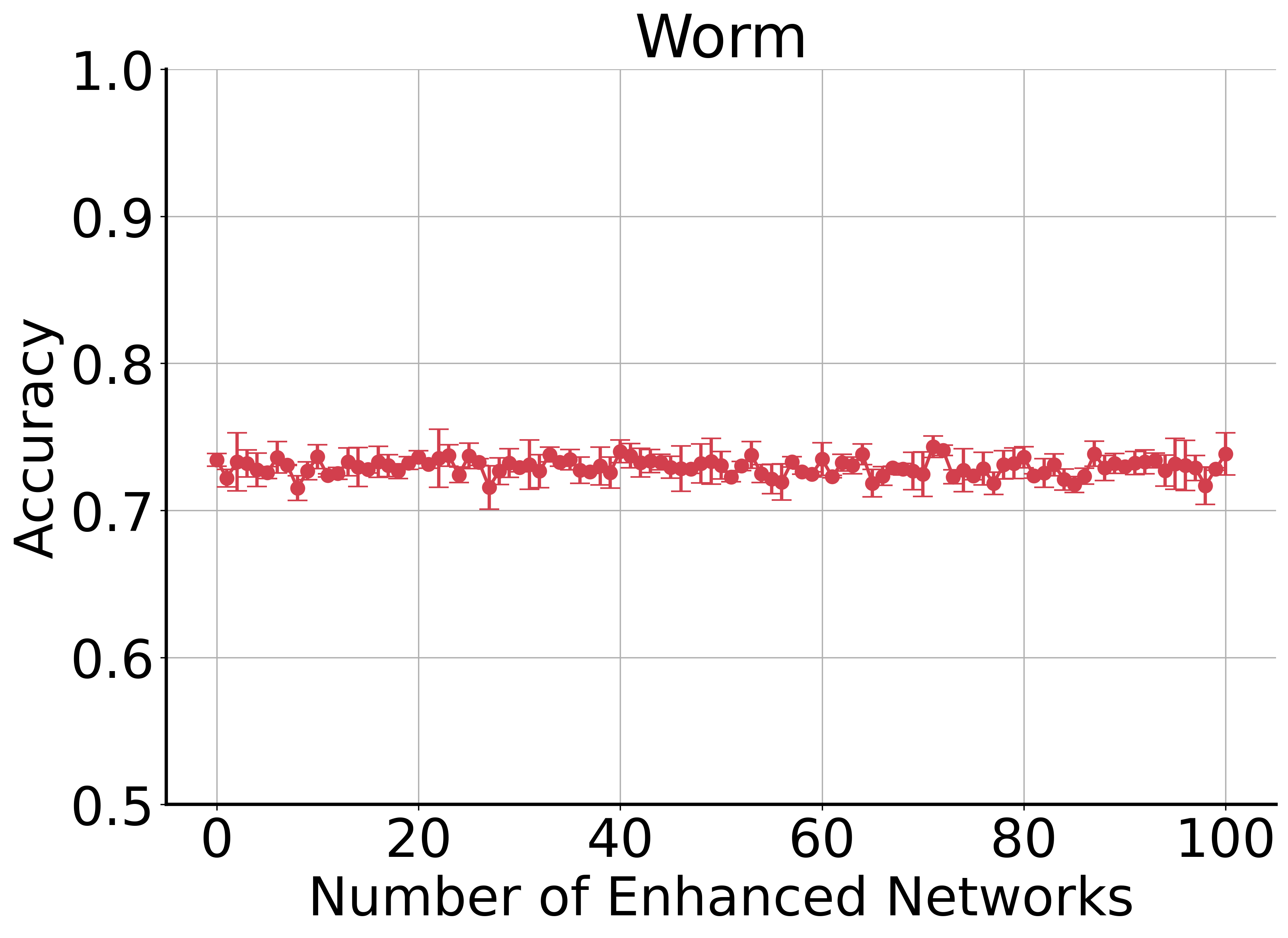}
		\caption*{} 
	\end{subfigure}
	\vspace{-0.5cm}
\end{figure*}
\begin{table*}[ht]
	\centering
	\caption{The upper section presents the results of joint training across different network combinations, while the lower section shows the additional accuracy improvements achieved by incorporating data augmentation into the joint training process. The table highlights performance across networks, including improvements on five previously unseen networks.}
	\label{tab:main_results}
	\resizebox{0.9\textwidth}{!}{
		\begin{tabular}{lccccccccccr}
			\toprule
			Train / Test & Airplane & Fruit & Thermody & Weaver & Worm & Bacteria & Coach & Fungi & Human & WTW & \textbf{Average} \\
			\midrule
			Average (Split) & 0.7641 & 0.7245 & \underline{0.5513} & 0.6696 & 0.6514 & \textit{0.7264} & \textit{0.8462} & \textit{0.7104} & \textit{0.6902} & \textit{\underline{0.6417}} & 0.6978 \\
			$\mathbb{AF}$ & 0.9828 & 0.8495 & \underline{0.4957} & 0.8201 & \underline{0.7020} & 0.9197 & 0.8663 & 0.8938 & 0.7438 & 0.8245 & 0.8098 \\
			$\mathbb{AFT}$ & 0.9761 & 0.8290 & 0.7420 & 0.7536 & \underline{0.6119} & 0.8940 & 0.8971 & 0.8246 & \underline{0.7235} & 0.7880 & 0.8030 \\
			$\mathbb{AFTW\mathrm{e}}$ & 0.9772 & 0.8112 & \underline{0.7218} & 0.8350 & \underline{0.6579} & 0.8205 & 0.8034 & 0.8573 & 0.7711 & 0.8098 & 0.8065 \\
			$\mathbb{AFW\mathrm{e}W\mathrm{o}}$ & 0.9775 & 0.8336 & \underline{0.4924} & 0.8452 & \underline{0.7352} & 0.7656 & 0.9124 & 0.8654 & 0.7602 & 0.7729 & 0.7960 \\
			$\mathbb{AFTW\mathrm{e}W\mathrm{o}}$ & 0.9772 & 0.8119 & \underline{0.7030} & 0.8346 & \underline{0.7329} & 0.8642 & 0.9133 & 0.8655 & 0.7551 & 0.8133 & 0.8271 \\
			\bottomrule
		\end{tabular}
	}
	
	\vspace{0.1cm}
	\resizebox{0.9\textwidth}{!}{
		\begin{tabular}{lcccccc}
			\toprule
			Train / Test & Bacteria & Coach & Fungi & Human & WTW & \textbf{Average} \\
			\midrule			
			$\mathbb{AFTWW}$ & 0.8642 (18.97+\%) & 0.9133 (7.93+\%) & 0.8655 (+21.83\%) & 0.7551 (+9.40\%) & 0.8133 (+26.74\%) & +16.98\%\\
			$\mathbb{AFTWW}$+$E_\mathbb{A}$ & 0.8873 (+22.15\%) & 0.9416 (+11.27\%) & 0.8956 (+26.07\%) & 0.7823 (+13.34\%) & 0.8172 (+27.35\%) & +20.04\%\\
			$\mathbb{AFTWW}$+$E_\mathbb{F}$ & 0.9213 (+26.83\%) & 0.9554 (+12.90\%) & 0.9128 (+28.49\%) & 0.7875 (+14.10\%) & 0.8211 (+27.96\%) & +22.06\%\\
			$\mathbb{AFTWW}$+$E_\mathbb{W\mathrm{o}}$ & 0.9336 (+28.52\%) & 0.9538 (+12.72\%) & 0.9145 (+28.73\%) & 0.7845 (+13.66\%) & 0.8249 (+28.55\%) & +22.44\%\\
			$\mathbb{AFTWW}$+$E_\mathbb{AF}$ & 0.9360 (+28.85\%) & 0.9609 (+13.55\%) & 0.9116 (+28.32\%) & 0.7837 (+13.55\%) & 0.7895 (+23.03\%) & +21.46\%\\
			$\mathbb{AFTWW}$+$E_\mathbb{FW\mathrm{o}}$ & 0.9138 (+25.80\%) & 0.9610 (+13.57\%) & 0.9101 (+28.11\%) & 0.7882 (+14.20\%) & 0.8158 (+27.13\%) & +21.76\%\\
			$\mathbb{AFTWW}$+$E_\mathbb{AFW\mathrm{o}}$ & 0.9063 (+24.77\%) & 0.9596 (+13.40\%) & 0.9071 (+27.69\%) & 0.7953 (+15.23\%) & 0.8159 (+27.15\%) & +21.65\%\\
			\bottomrule
		\end{tabular}
	}
\end{table*}
\subsection{Enhancing Model Performance through Joint Training and Data Augmentation}
\subsubsection{Impact of Joint Training}\label{exp:multi}
Table~\ref{tab:main_results} provides a detailed comparison of model performance under different training scenarios, revealing the benefits of joint training. The first row of the table, labeled as Average (Split), shows the average accuracy of models trained on each individual network and tested across all networks. This serves as a baseline for understanding how single-network training performs in isolation.

Rows labeled $\mathbb{AF}$ through $\mathbb{AFTW\mathrm{e}W\mathrm{o}}$ represent models trained on progressively larger combinations of networks. 
$\mathbb{A}$, $\mathbb{F}$, $\mathbb{T}$, $\mathbb{W\mathrm{e}}$, and $\mathbb{W\mathrm{o}}$ represent networks Airplane, Fruit, Thermody, Weaver, and Worm, respectively. $\mathbb{AF}$ denotes the union of networks Airplane and Fruit.
As more networks are incorporated, the model's performance becomes increasingly stable, with the worst-case accuracy improving significantly. For instance, the worst accuracy of $\mathbb{AF}$ is 0.4957, whereas $\mathbb{AFTW\mathrm{e}W\mathrm{o}}$ achieves 0.7030. This indicates that adding networks reduces variability and enhances robustness across test scenarios.
On average, the fused training $\mathbb{AFTW\mathrm{e}W\mathrm{o}}$ achieves 0.8133, demonstrating an improvement over the Average (Split) of 0.6978 from separate training.

Interestingly, the addition of seemingly unrelated networks, such as Thermody, not only improves accuracy but also stabilizes the model. As shown in Table~\ref{tab:network_performance}, when Thermody is trained in isolation, its predictions on other networks consistently exhibit poor performance, with all accuracies below 0.6. However, comparing the results of $\mathbb{AFW\mathrm{e}W\mathrm{o}}$ and $\mathbb{AFTW\mathrm{e}W\mathrm{o}}$, incorporating Thermody into the training process enhances predictions on Bacteria and WTW. This highlights the complementary value of diverse networks, demonstrating that even those with poor standalone transferability can contribute critical information to improve both model stability and overall performance when included in joint training.

Overall, joint training proves to be a powerful approach, leveraging network diversity to achieve robust and accurate predictions. This strategy mitigates the limitations of single-network training and underscores the importance of comprehensive data integration in predictive tasks.

\begin{figure*}[htbp]
	\centering
	\caption{Performance on Unseen Networks with Augmentation}
	\label{fig:augmentation_unseen}
	\begin{subfigure}{0.32\textwidth}
		\centering
		\includegraphics[width=\linewidth]{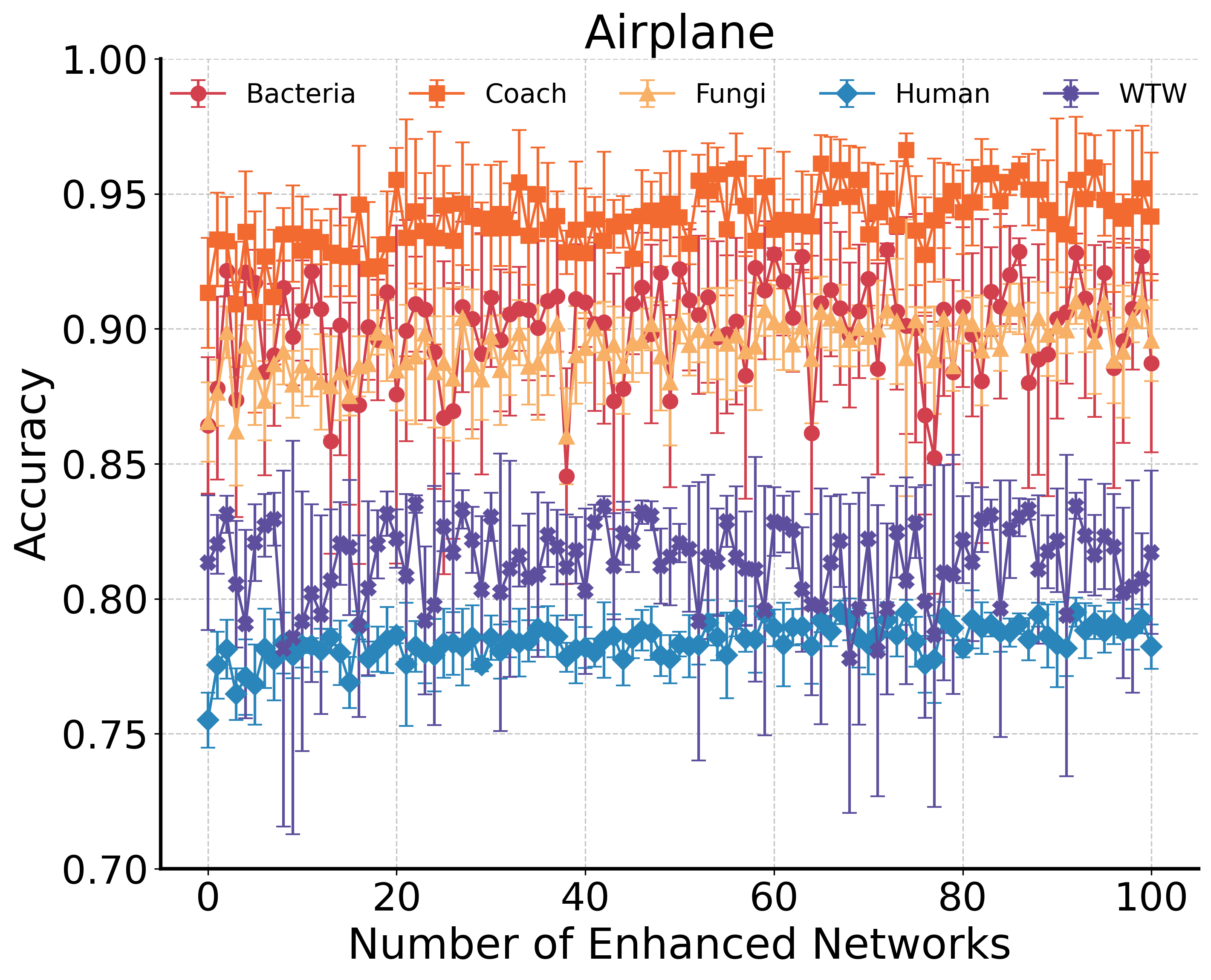}
		\caption*{} 
	\end{subfigure}
	\begin{subfigure}{0.32\textwidth}
		\centering
		\includegraphics[width=\linewidth]{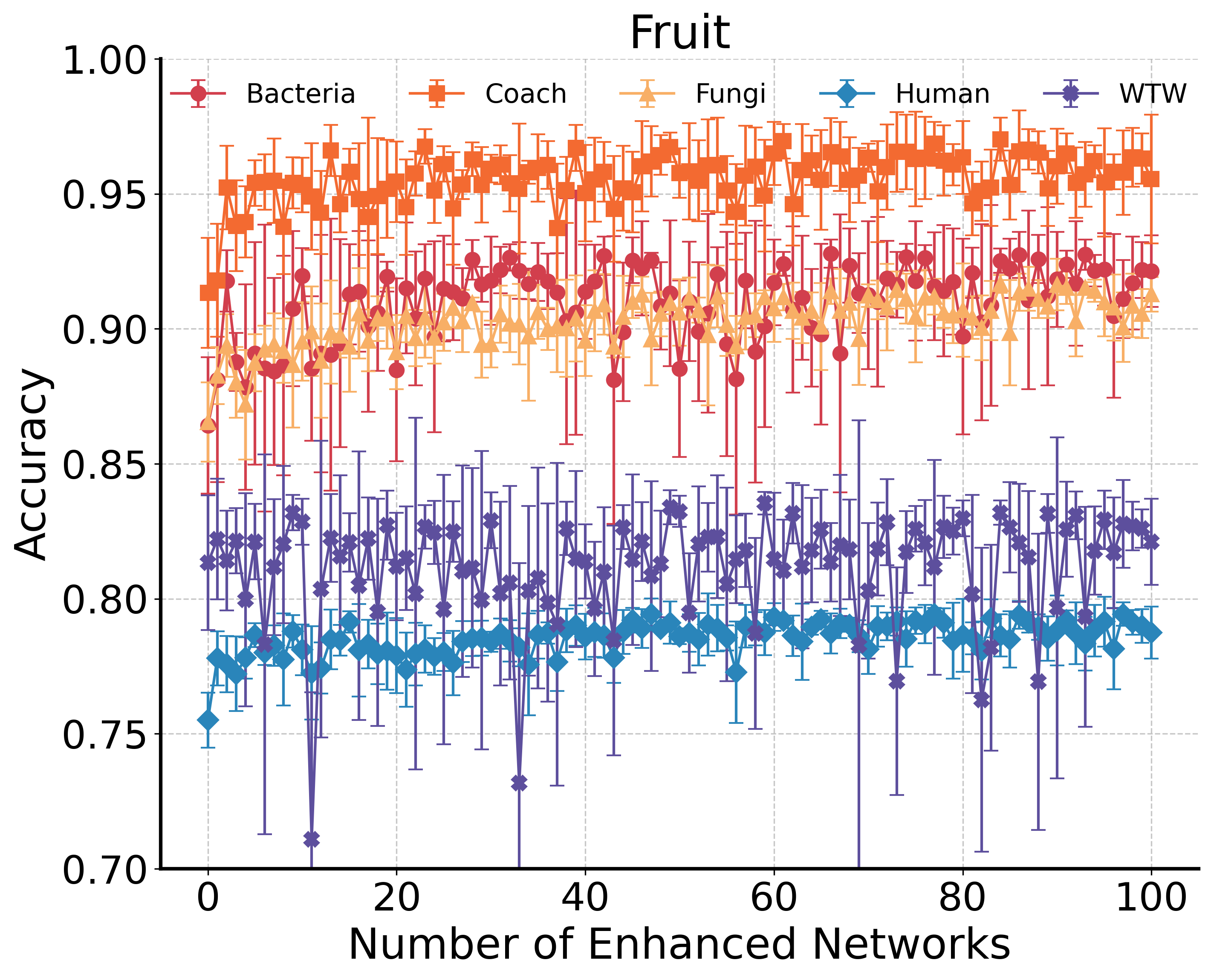}
		\caption*{} 
	\end{subfigure} 
	\begin{subfigure}{0.32\textwidth}
		\centering
		\includegraphics[width=\linewidth]{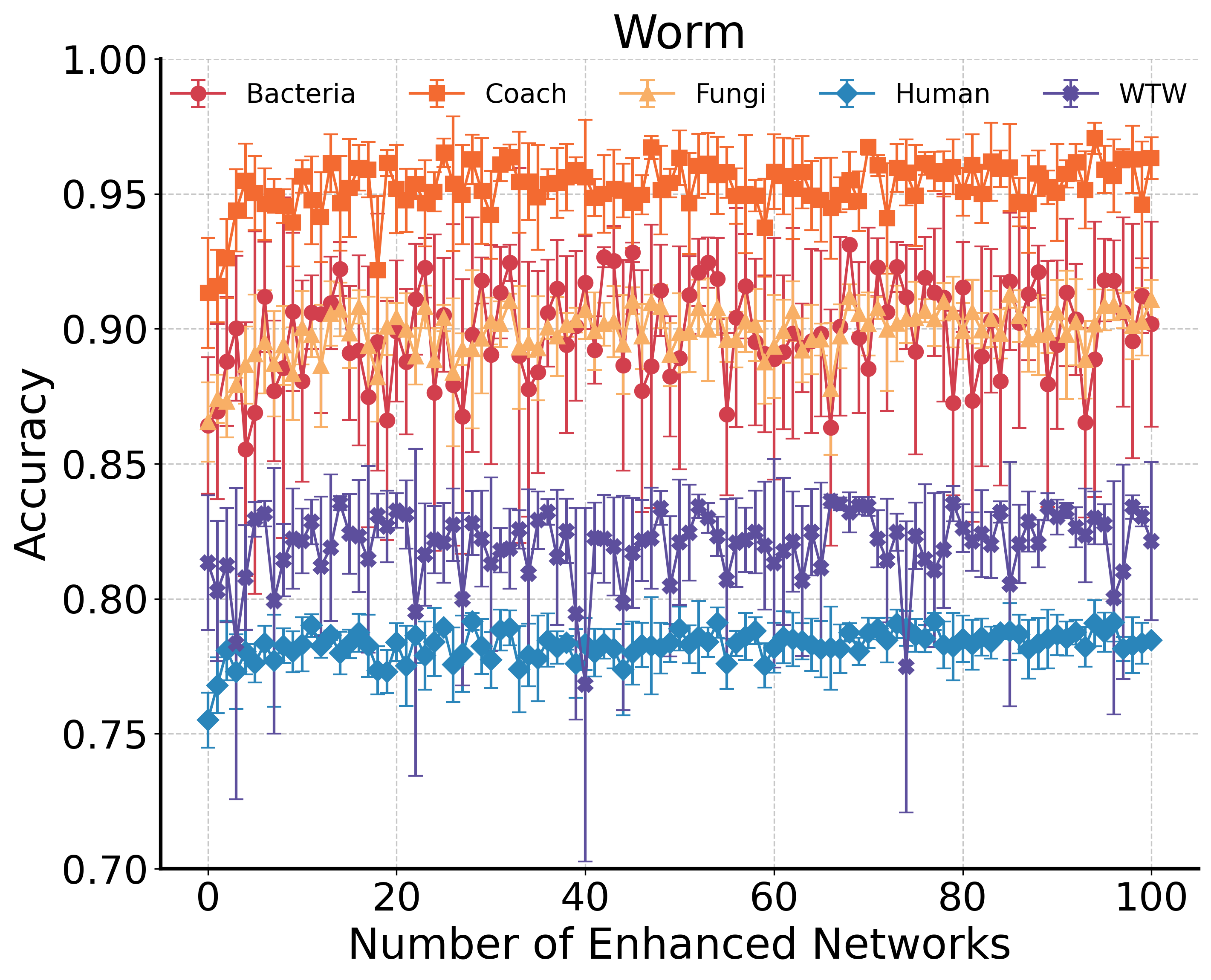}
		\caption*{} 
	\end{subfigure}\\ \vspace{-18pt}
	
	\begin{subfigure}{0.32\textwidth}
		\centering
		\includegraphics[width=\linewidth]{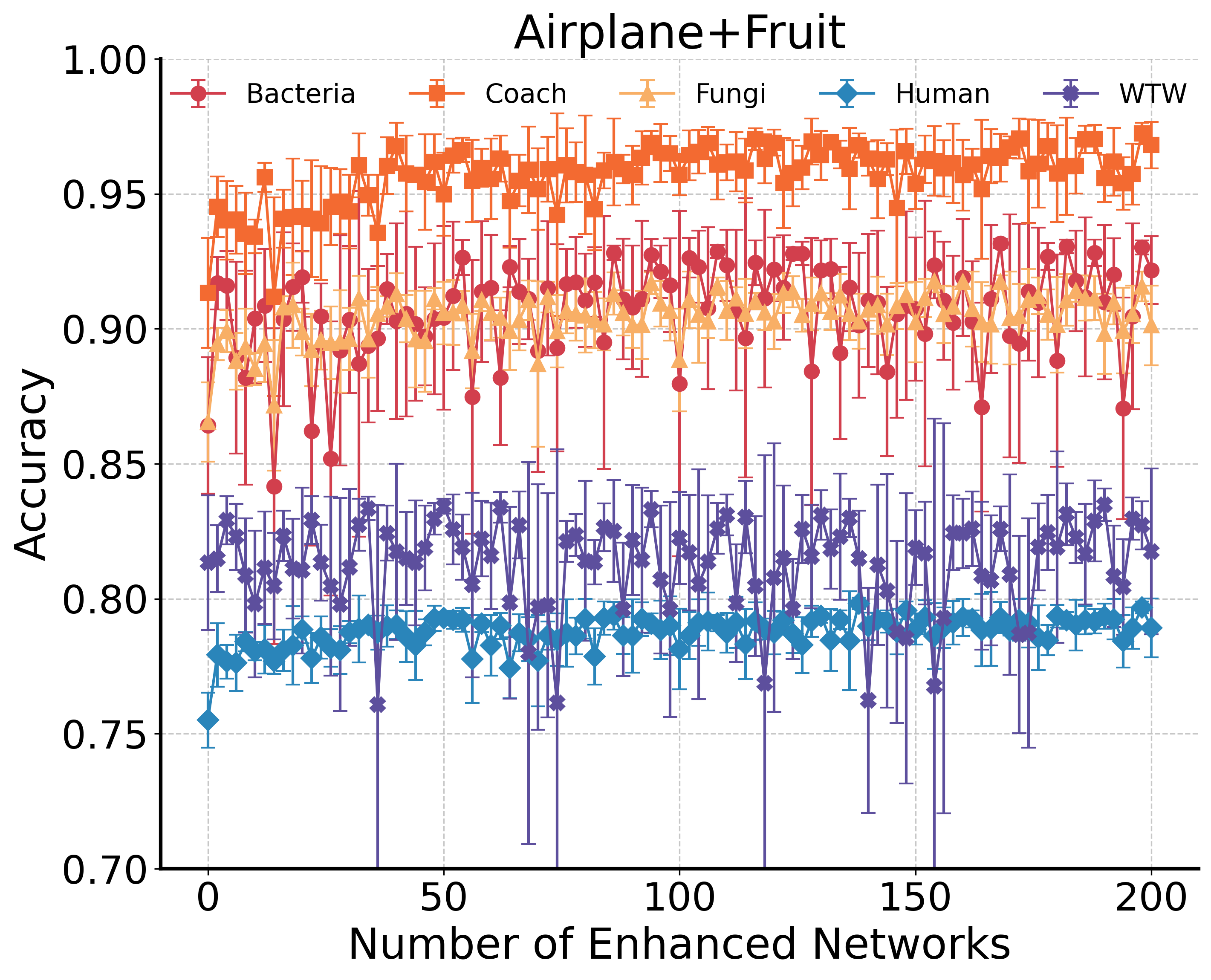}
		\caption*{} 
	\end{subfigure} 
	\begin{subfigure}{0.32\textwidth}
		\centering
		\includegraphics[width=\linewidth]{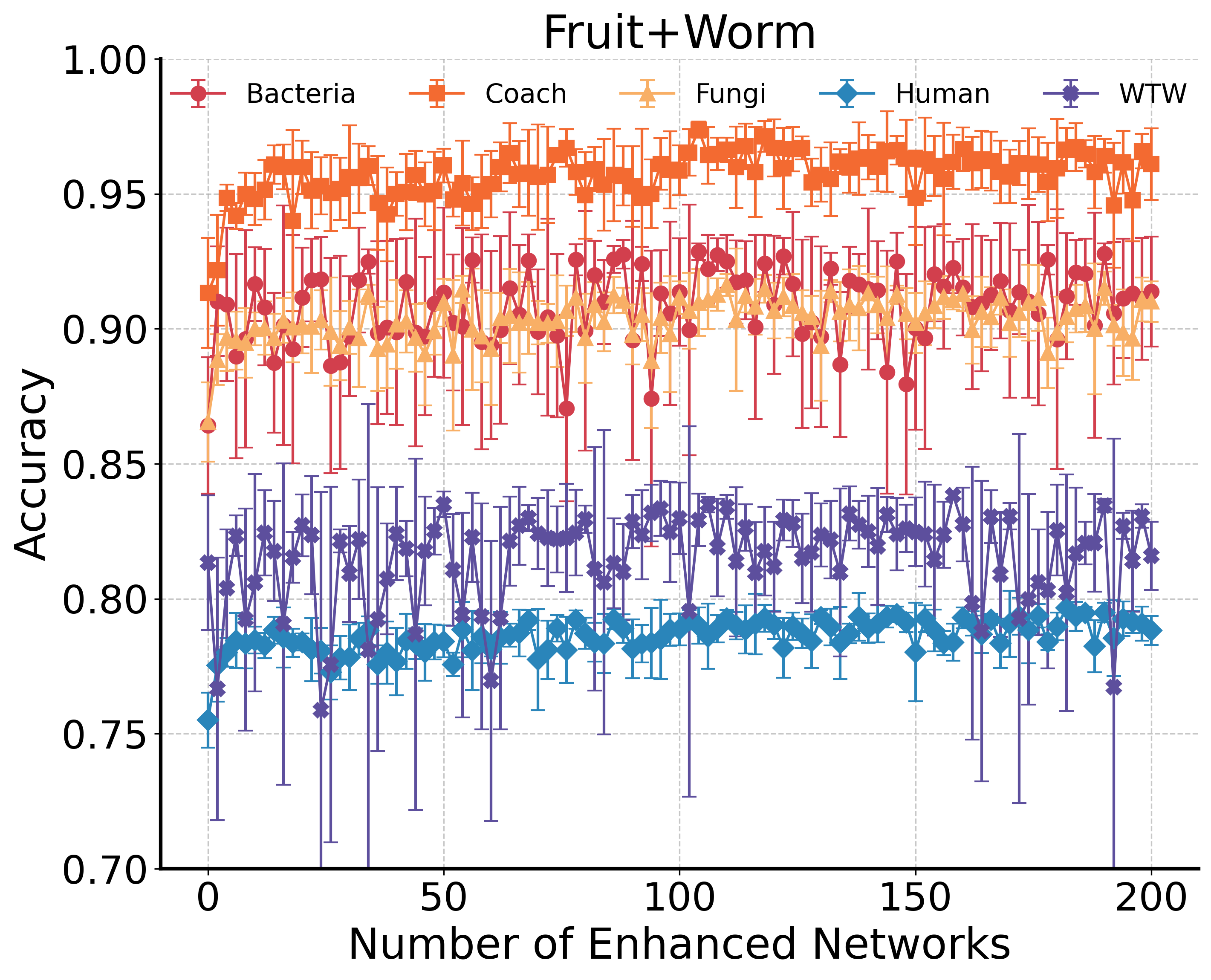}
		\caption*{} 
	\end{subfigure}
	\begin{subfigure}{0.32\textwidth}
		\centering
		\includegraphics[width=\linewidth]{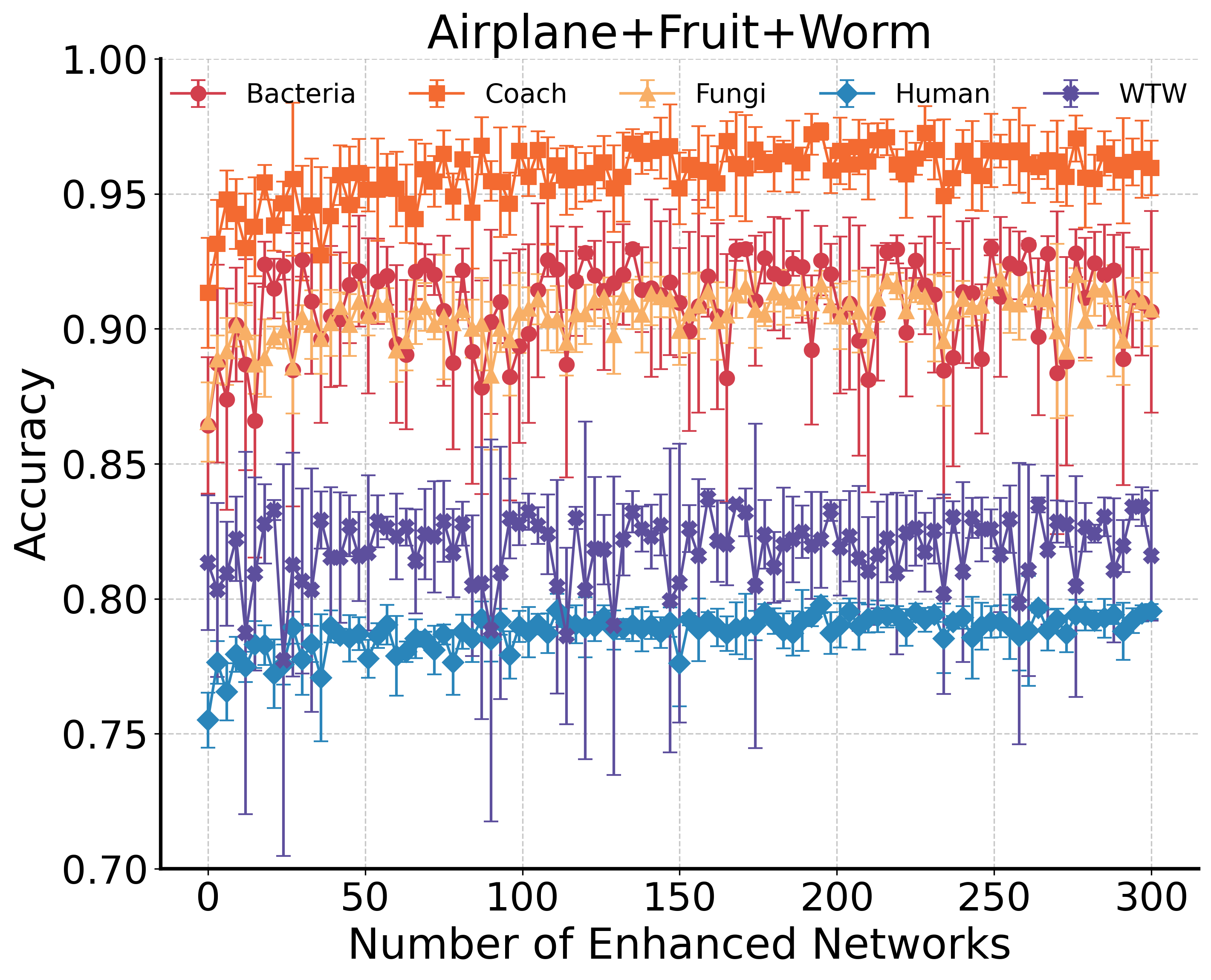}
		\caption*{} 
	\end{subfigure}	\\ \vspace{-22pt}
\end{figure*}

\subsubsection{Benefits of Data Augmentation}
While joint training has proven to enhance model accuracy, particularly on unseen networks, we sought to explore further improvements through the use of additional training samples. However, obtaining networks with edge generation timestamps is a labor-intensive and resource-consuming process. To overcome this limitation, we employed generative methods to create synthetic networks, enabling us to generate an arbitrary number of samples without the acquisition difficulties associated with real-world data.

The experimental results, as shown in Fig. \ref{fig:augmentation_self}, reveal that adding generated networks does not improve performance when predicting the withheld parts of the training networks themselves. This suggests that when sufficient real edges are already present, adding synthetic edges provides no additional advantage. Importantly, the performance does not degrade either, remaining consistent despite the augmented training data. 
In Table \ref{tab:main_results}, the results on the temporal networks Airplane, Fruit, Thermody, Weaver, and Worm, which are included in the training set, are consistent with those of the $\mathbb{AFTW\mathrm{e}W\mathrm{o}}$. Therefore, we just provide a detailed presentation of the performance on unseen static networks.
Each training network generates 100 corresponding augmented networks. During integrated training, only the edges absent in the original training network are retained for training.

The benefits of data augmentation become apparent when evaluating the model's accuracy on unseen networks. Models trained with augmented samples from Airplane, Fruit, and Worm demonstrated significant accuracy improvements on previously unseen networks. However, augmenting samples from Thermody did not yield similar gains. This is likely due to the inherent structural dissimilarity between Thermody and other networks, which may cause the model to deviate from generalizable patterns and overfit to the unique characteristics of Thermody.

To further leverage the advantages of augmentation, we combined the augmented samples from Airplane, Fruit, and Worm for training. The results, shown in Fig. \ref{fig:augmentation_unseen}, indicate that combining augmented samples leads to further performance improvements. Specifically, combining Airplane and Fruit, Fruit and Worm, and all three networks together, consistently improved accuracy. The improvement magnitudes are detailed in Fig. \ref{fig:augmentation_unseen}. The specific numerical results are presented in Table \ref{tab:main_results}, demonstrating the effectiveness of augmentation. 
Notably, as shown in Table \ref{tab:main_results}, fused training achieves a 16.98\% improvement, while data augmentation provides an additional 5.46\% boost, resulting in a total improvement of 22.44\%.
Additionally, to further validate the effectiveness of our augmentation strategy, we conducted experiments not only under the fused training model but also with single-network training, where one original network was paired with an augmented network. For details, please refer to Appendix \ref{apdx:enhance}.

These findings demonstrate the potential of data augmentation to enhance model performance, particularly on unseen networks, while highlighting the importance of selecting networks with complementary structures for augmentation to maximize its effectiveness.

\begin{figure*}[htbp]
	\centering
	\begin{subfigure}{0.3\textwidth}
		\centering
		\includegraphics[width=\linewidth]{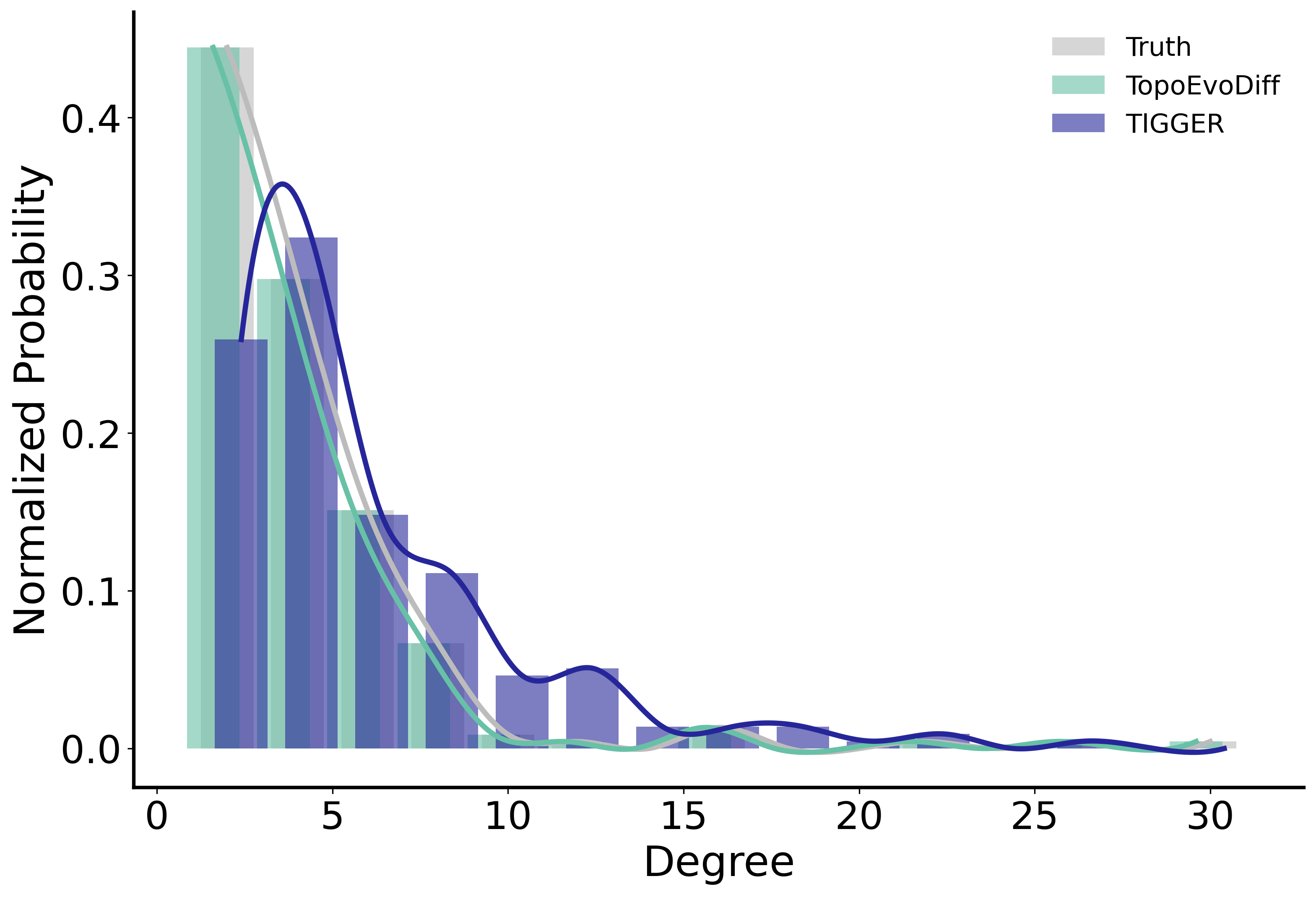}
		\caption*{} 
	\end{subfigure}
	\begin{subfigure}{0.3\textwidth}
		\centering
		\includegraphics[width=\linewidth]{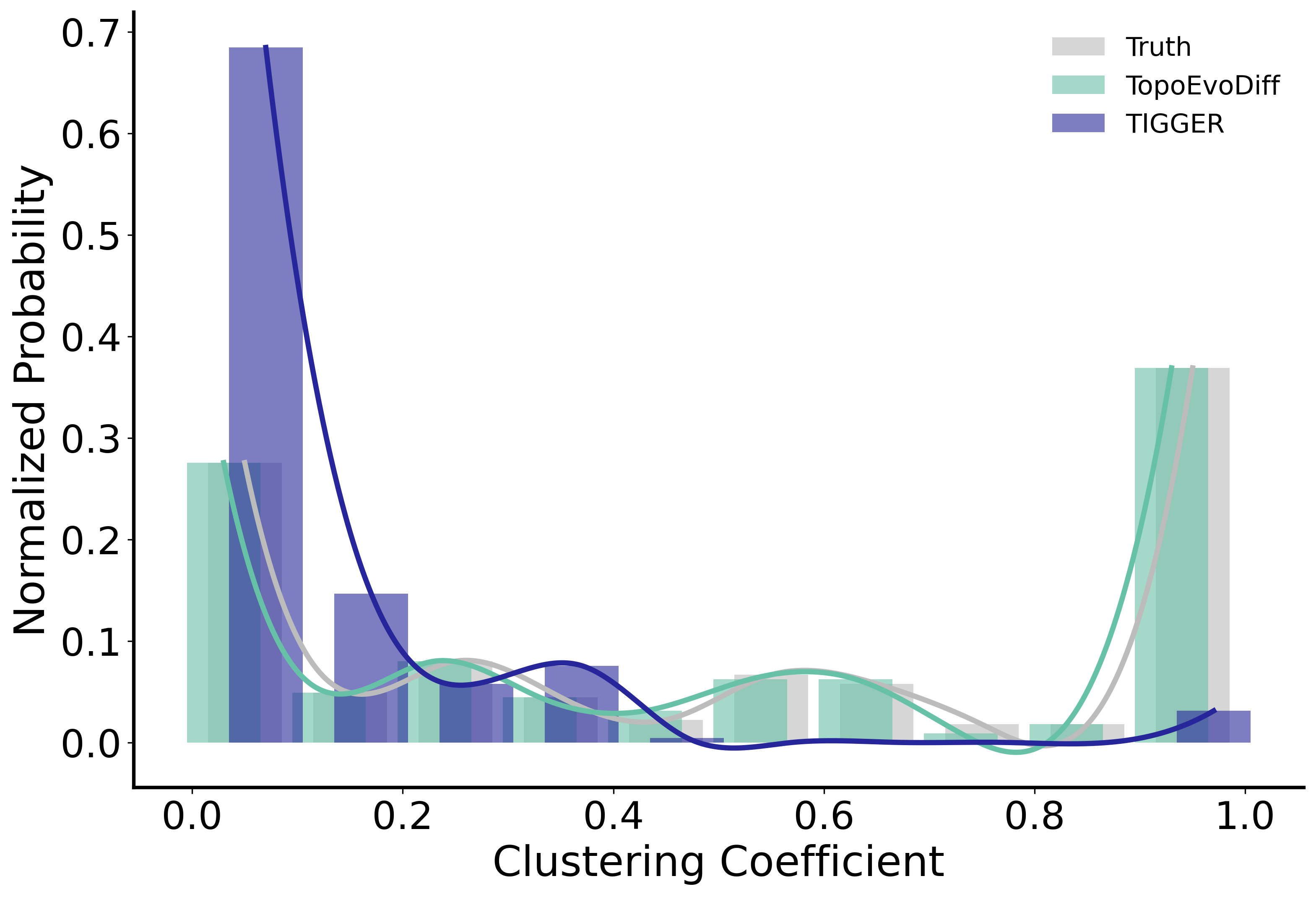}
		\caption*{} 
	\end{subfigure}
	\begin{subfigure}{0.3\textwidth}
		\centering
		\includegraphics[width=\linewidth]{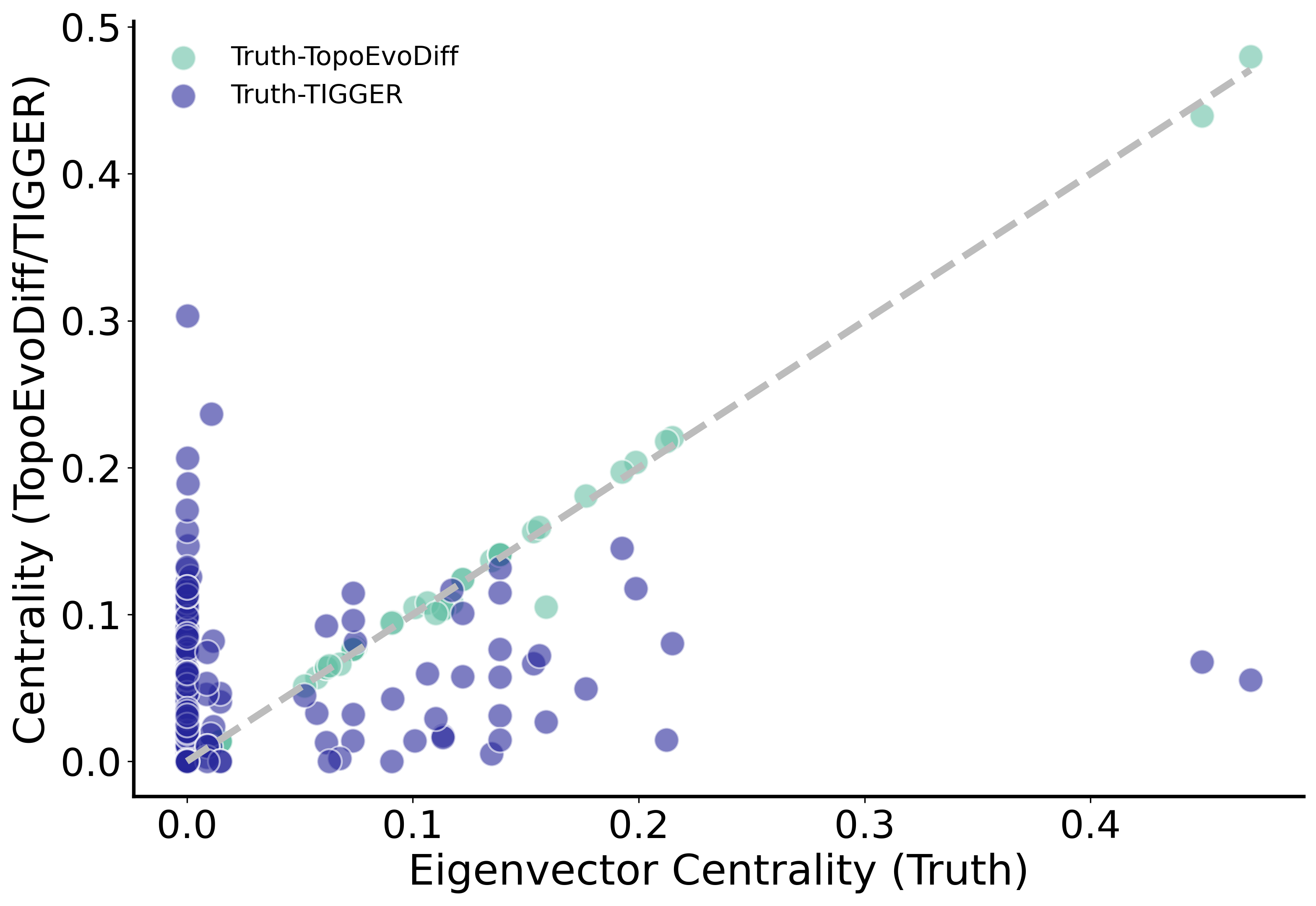}
		\caption*{} 
	\end{subfigure}
	\vspace{-15pt}
	\begin{subfigure}{0.3\textwidth}
		\centering
		\includegraphics[width=\linewidth]{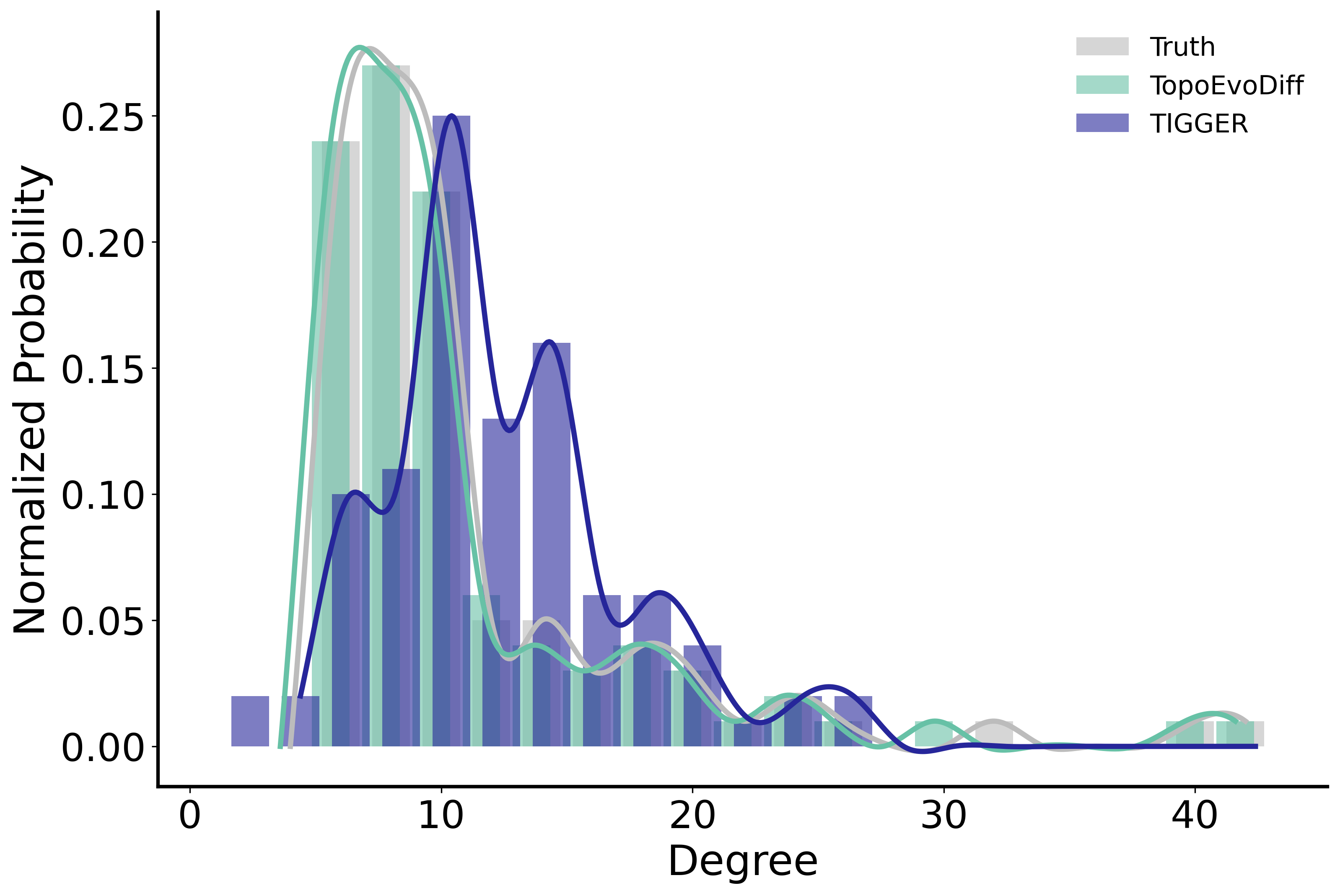}
		\caption*{} 
	\end{subfigure}
	\begin{subfigure}{0.3\textwidth}
		\centering
		\includegraphics[width=\linewidth]{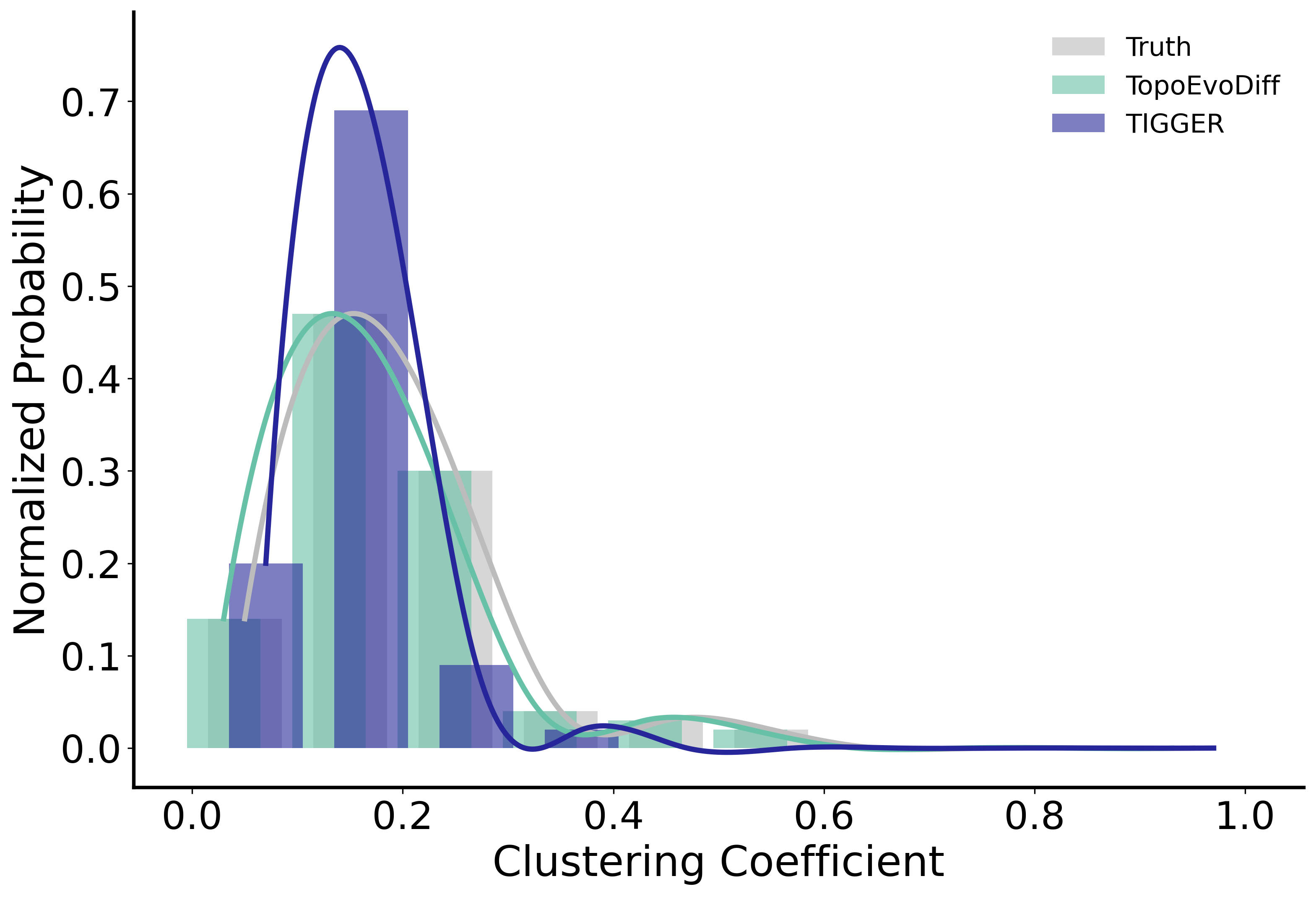}
		\caption*{} 
	\end{subfigure}
	\begin{subfigure}{0.3\textwidth}
		\centering
		\includegraphics[width=\linewidth]{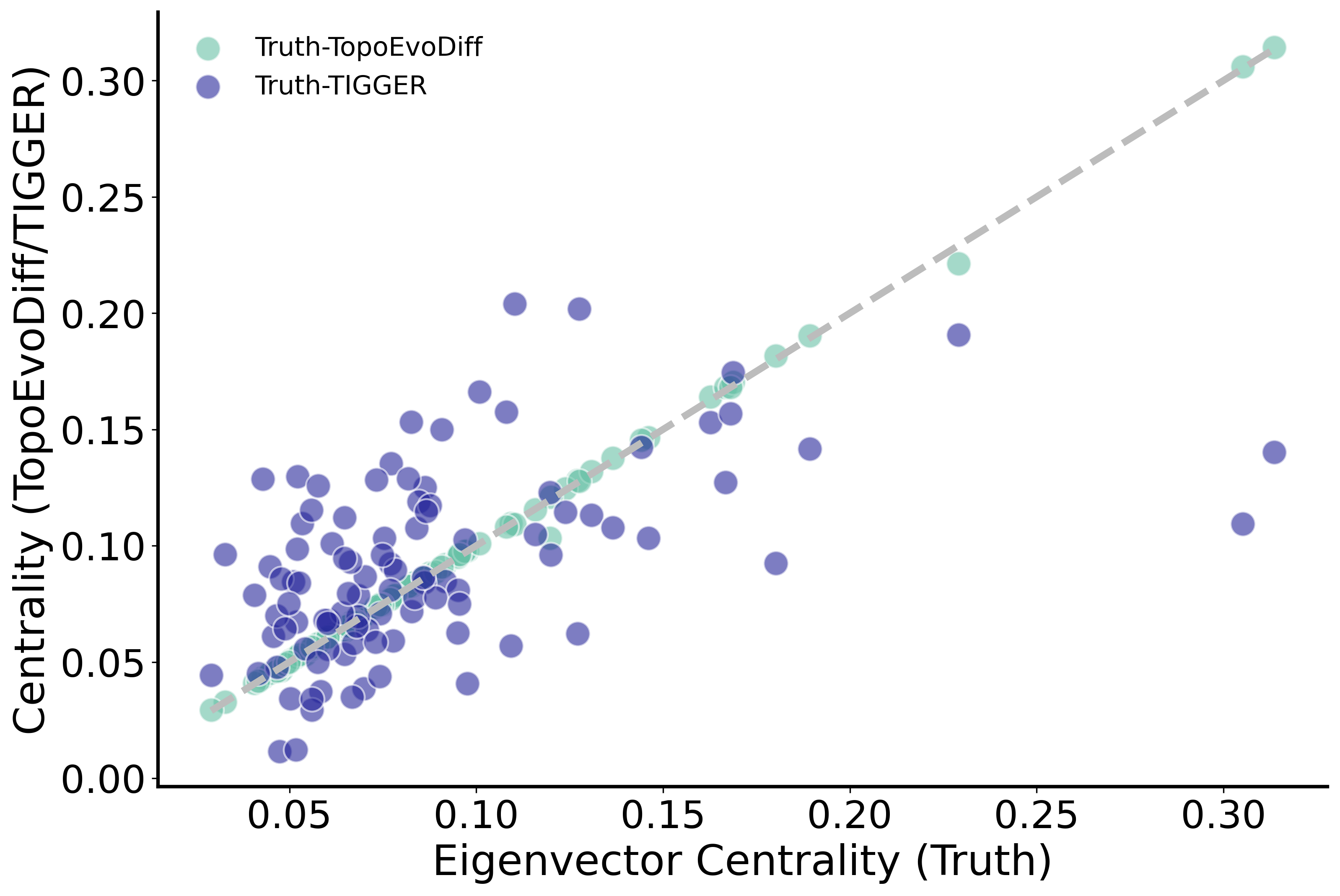}
		\caption*{} 
	\end{subfigure}
	\caption{Similarity between the generated networks and the original networks.}
	\label{fig:similarity}
\end{figure*}

\subsection{Performance of Baseline Models with Augmentation.}
\begin{table}[ht]
	\centering
	\caption{Performance of Augmented Baseline Models}
	\label{tab:transposed_network_performance}
	\resizebox{0.48\textwidth}{!}{
		\begin{tabular}{lccccc}
			\toprule
			Model / Network & Bacteria &  Coach &  Fungi &  Human &  WTW  \\
			\hline
			$\mathbb{AFTWW}$ &  0.8642 & 0.9133 & 0.8655 & 0.7551 & 0.8133 \\  
			RF & 0.8409 & 0.9203 & 0.8745 & 0.7645 & 0.8199  \\ 
			TIGGER & 0.9051 & 0.9402 & 0.8977 & 0.7745 & 0.8215 \\ 
			NetGAN & 0.8945 & 0.9121 & 0.8883 & 0.7708 & 0.8202 \\ 
			Digress & 0.9089 & 0.9345 & 0.9047 & 0.7754 & 0.8205 \\ 
			\textbf{TopoEvoDiff} & 0.9336 & 0.9538 & 0.9145 & 0.7845 & 0.8249 \\ 
			\bottomrule
		\end{tabular}
	}
\end{table}

To further validate the effectiveness of our augmentation strategy, we applied it to networks generated using alternative baseline methods and compared the results. Specifically, synthetic networks were generated through baseline approaches, and these augmented samples were then combined with the original training data for joint training. The results revealed that, although the augmentation increased the diversity of the training samples, the performance improvements achieved by the baseline methods were limited compared to those observed with our proposed approach.

In this experiment, five real networks were used to train the generation models, and each trained model generated 100 augmented networks. To maximize the performance gains, we selected the augmented network set from the five generated options that brought the most significant improvement when used for augmentation. Among these, the Worm network generated by our method led to the largest performance enhancement.
The augmented baseline models showed some performance improvement compared to their non-augmented counterparts. However, the magnitude of the improvement was notably smaller than that achieved with our proposed method. Across all test networks, the accuracy showed modest gains, indicating that while the additional synthetic samples did contribute to enhancing the predictive capabilities of the baseline models, the improvements were not as significant as those observed with our augmentation strategy. This suggests that, although the baseline methods benefited from augmentation, they were less effective in capturing the structural and temporal dynamics of the networks, resulting in smaller performance gains compared to our approach.

\subsection{Comparison of Generated Network Quality}
\begin{table}[ht]
	\centering
	\caption{NRMSE comparison of different models across five networks. TopoEvoDiff shows consistent improvements over the second-best results (underlined).}
	\label{tab:performance_nrmse}
	\resizebox{0.48\textwidth}{!}{
		\begin{tabular}{lccccc}
			\toprule
			Model / Network & Airplane &  Fruit &  Thermody &  Weaver &  Worm  \\
			\hline
			RF & 0.6547 & 0.6521 & 0.5901 & 0.7504 & 0.6179 \\ 
			TIGGER & \underline{0.6398} & 0.6348 & \underline{0.5674} & 0.7311 & 0.6039 \\ 
			NetGAN & 0.6478 & 0.6413 & 0.5802 & 0.7402 & 0.6085 \\ 
			Digress & 0.6403 & \underline{0.6294} & 0.5758 & \underline{0.7258} & \underline{0.6038} \\ \hline
			\textbf{TopoEvoDiff} & \makecell{0.5814 \\ (+9.13\%)} & \makecell{0.5408 \\ (+14.07\%)} & \makecell{0.5540 \\ (+2.37\%)} & \makecell{0.6863 \\ (+5.44\%)} & \makecell{0.5605 \\ (+7.17\%)} \\ 
			\bottomrule
		\end{tabular}
	}
\end{table}

To evaluate the effectiveness of our augmentation method compared to baseline approaches, we analyzed the discrepancy between generated networks and their corresponding real networks. The primary metric used for this evaluation was the Normalized Root Mean Square Error (NRMSE), which measures the difference between the generated edge generation times and the real timestamps.

Table \ref{tab:performance_nrmse} presents the NRMSE results for networks generated using our method and the baseline approaches. Our method, which learns directly from real networks, consistently achieves lower NRMSE values compared to the baselines. This indicates that the generated networks more accurately replicate the temporal dynamics of edge generation observed in the real data.
Baseline methods exhibit higher NRMSE values, suggesting a lack of fidelity in capturing the nuances of real network evolution. The inability to closely match the temporal properties of real networks limits the utility of baseline-generated samples in enhancing model performance.

\begin{figure*}[htbp]
	\centering
	\caption{In the task of predicting unknown edge generations within the same network, the training set investigates the relationship between the proportion of edges with known generation times and the prediction accuracy.}
	\label{fig:train_ratio}
	\begin{subfigure}{0.23\textwidth}
		\centering
		\includegraphics[width=\linewidth]{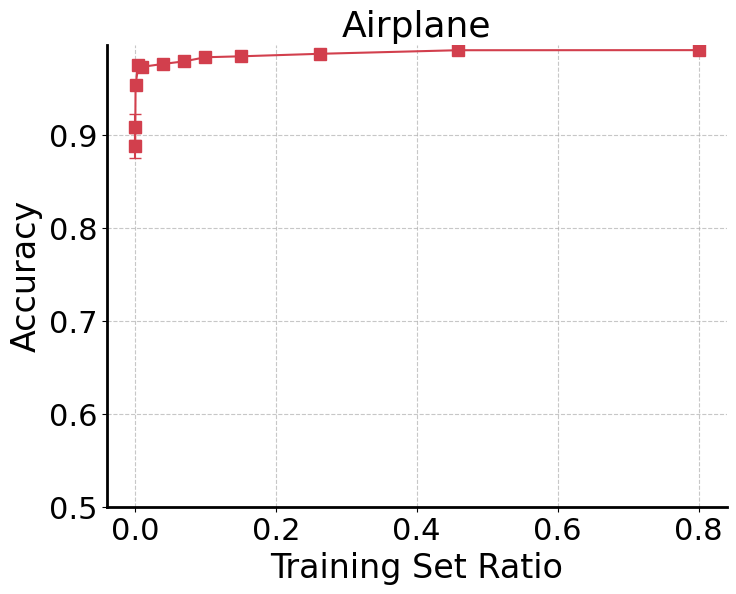}
		\caption*{} 
	\end{subfigure}
	\begin{subfigure}{0.23\textwidth}
		\centering
		\includegraphics[width=\linewidth]{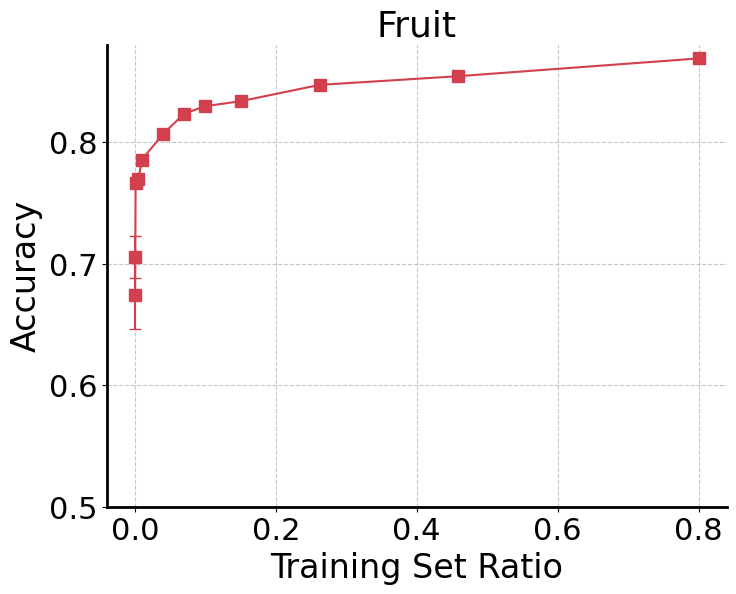}
		\caption*{} 
	\end{subfigure}
	\begin{subfigure}{0.23\textwidth}
		\centering
		\includegraphics[width=\linewidth]{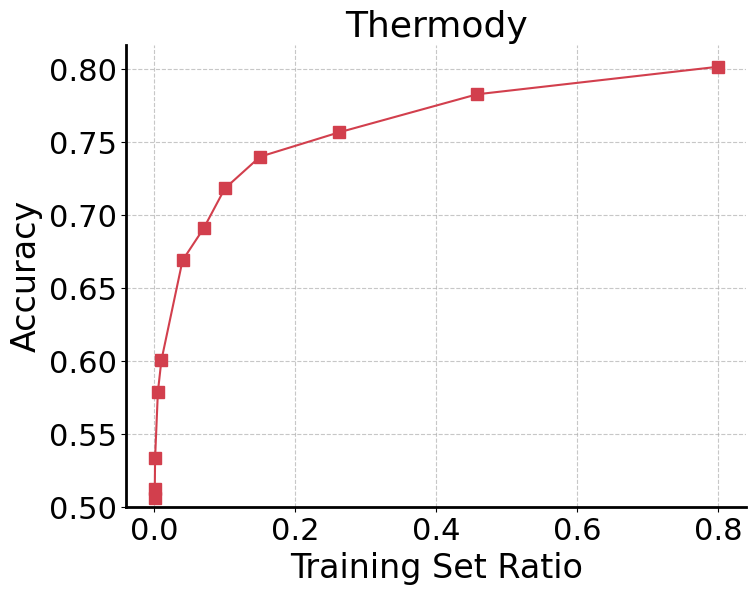}
		\caption*{} 
	\end{subfigure}
	\begin{subfigure}{0.23\textwidth}
		\centering
		\includegraphics[width=\linewidth]{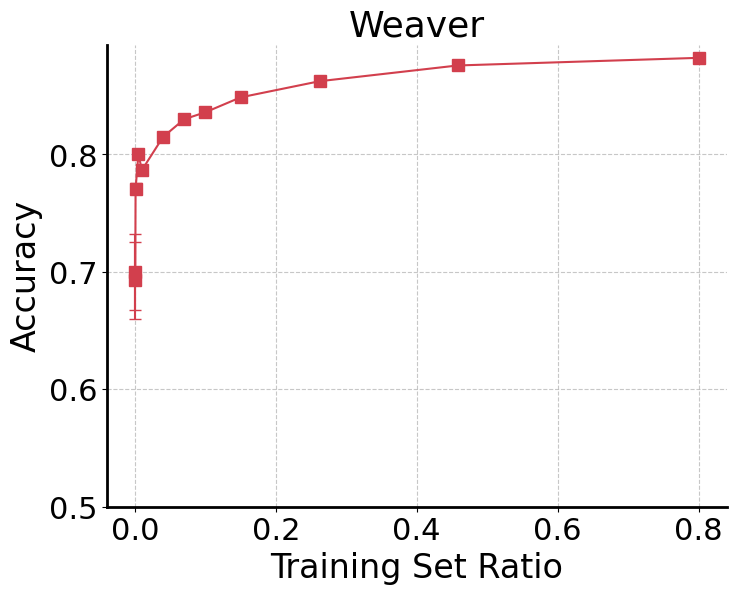}
		\caption*{} 
	\end{subfigure}	\\ \vspace{-15pt}
	\begin{subfigure}{0.23\textwidth}
		\centering
		\includegraphics[width=\linewidth]{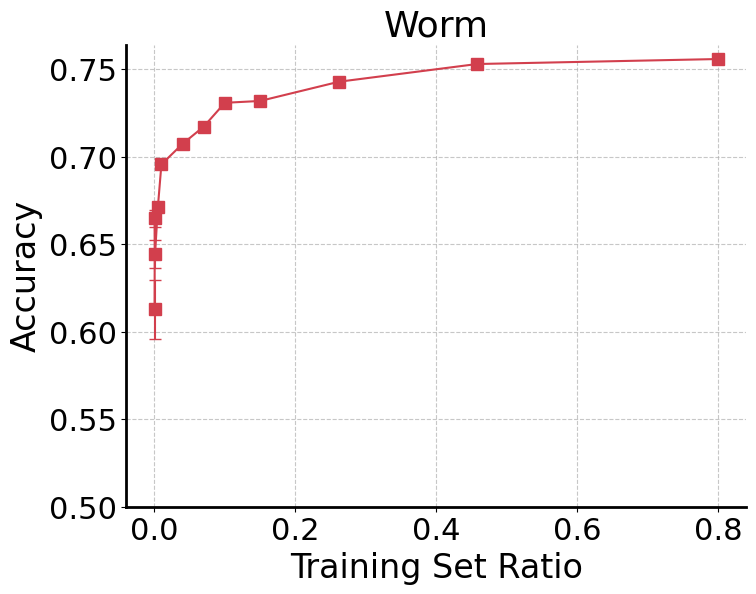}
		\caption*{} 
	\end{subfigure}
	\begin{subfigure}{0.23\textwidth}
		\centering
		\includegraphics[width=\linewidth]{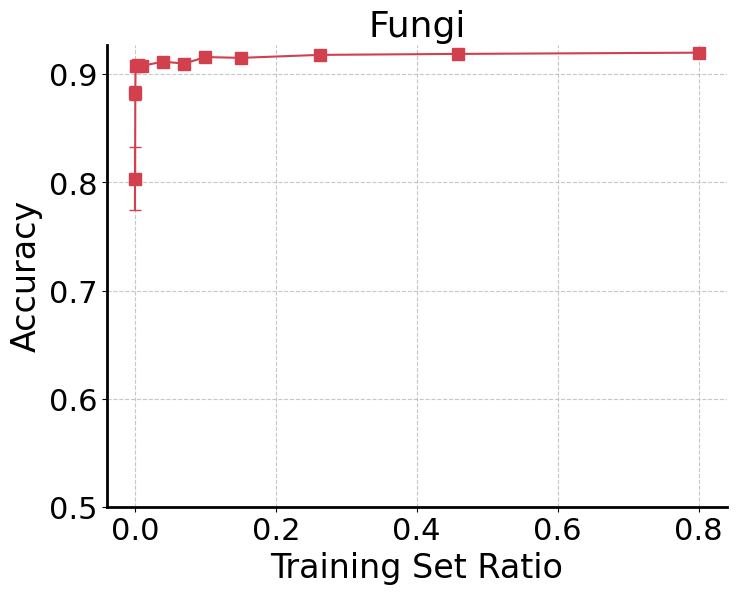}
		\caption*{} 
	\end{subfigure}
	\begin{subfigure}{0.23\textwidth}
		\centering
		\includegraphics[width=\linewidth]{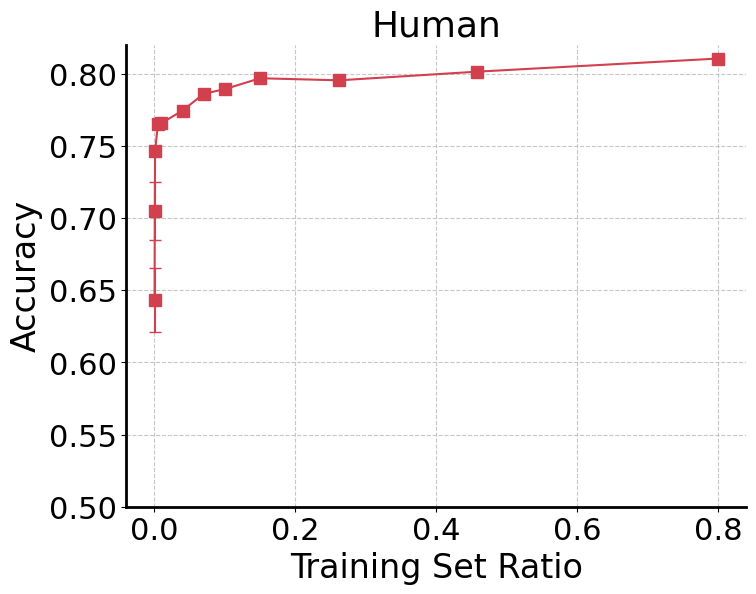}
		\caption*{} 
	\end{subfigure}
	\begin{subfigure}{0.23\textwidth}
		\centering
		\includegraphics[width=\linewidth]{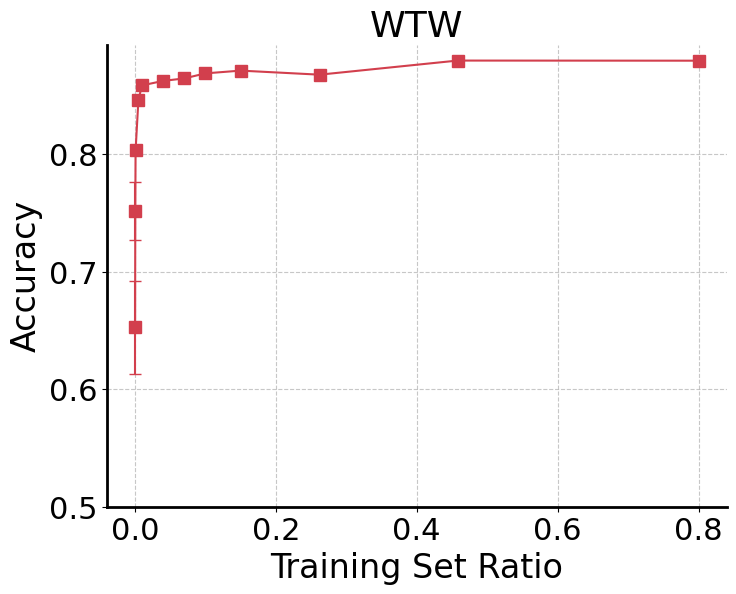}
		\caption*{} 
	\end{subfigure} \vspace{-15pt}
\end{figure*}

\begin{figure}[htbp]
	\centering
	\caption{Visualization of the differences in network features and edge generation time distributions across networks.}
	\label{fig:benefit}
	\begin{subfigure}{0.46\textwidth}
		\centering
		\includegraphics[width=\linewidth]{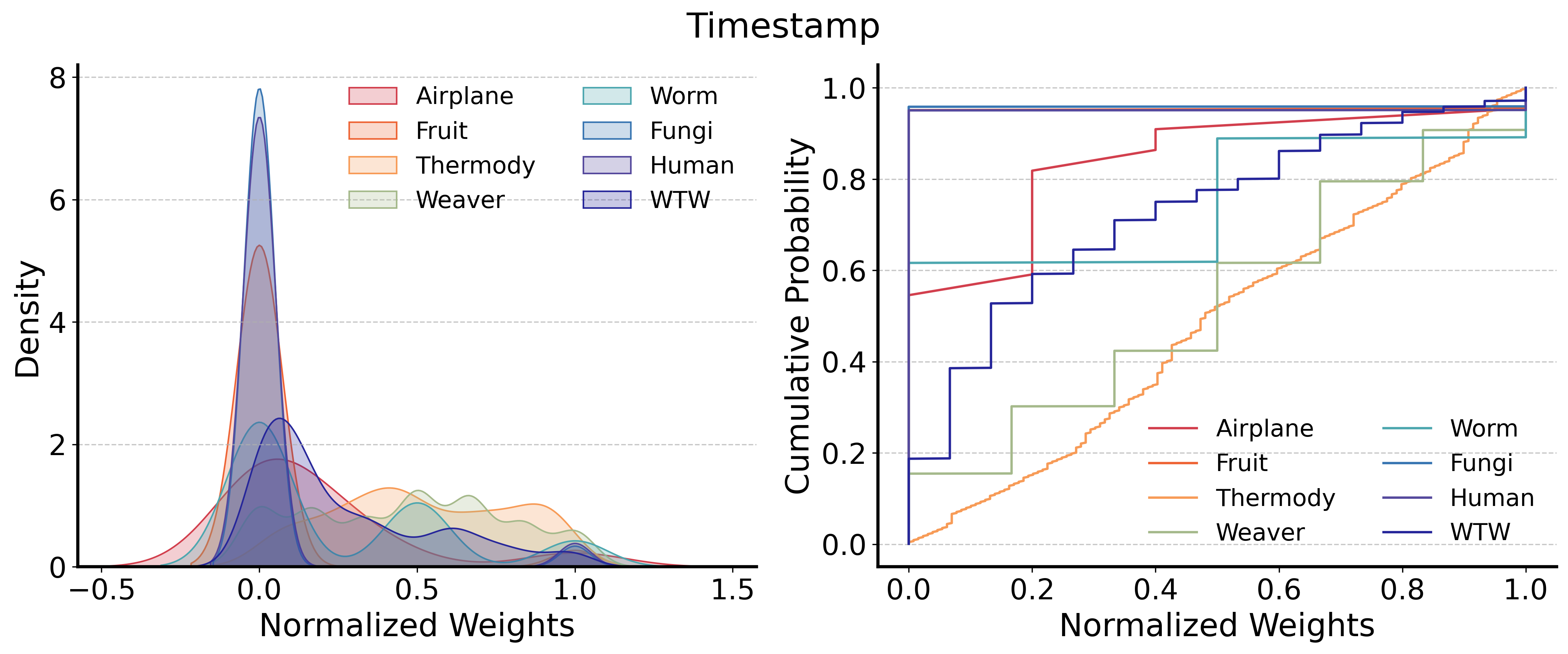}
		\caption*{} 
	\end{subfigure}\vspace{-18pt}
	\begin{subfigure}{0.46\textwidth}
		\centering
		\includegraphics[width=\linewidth]{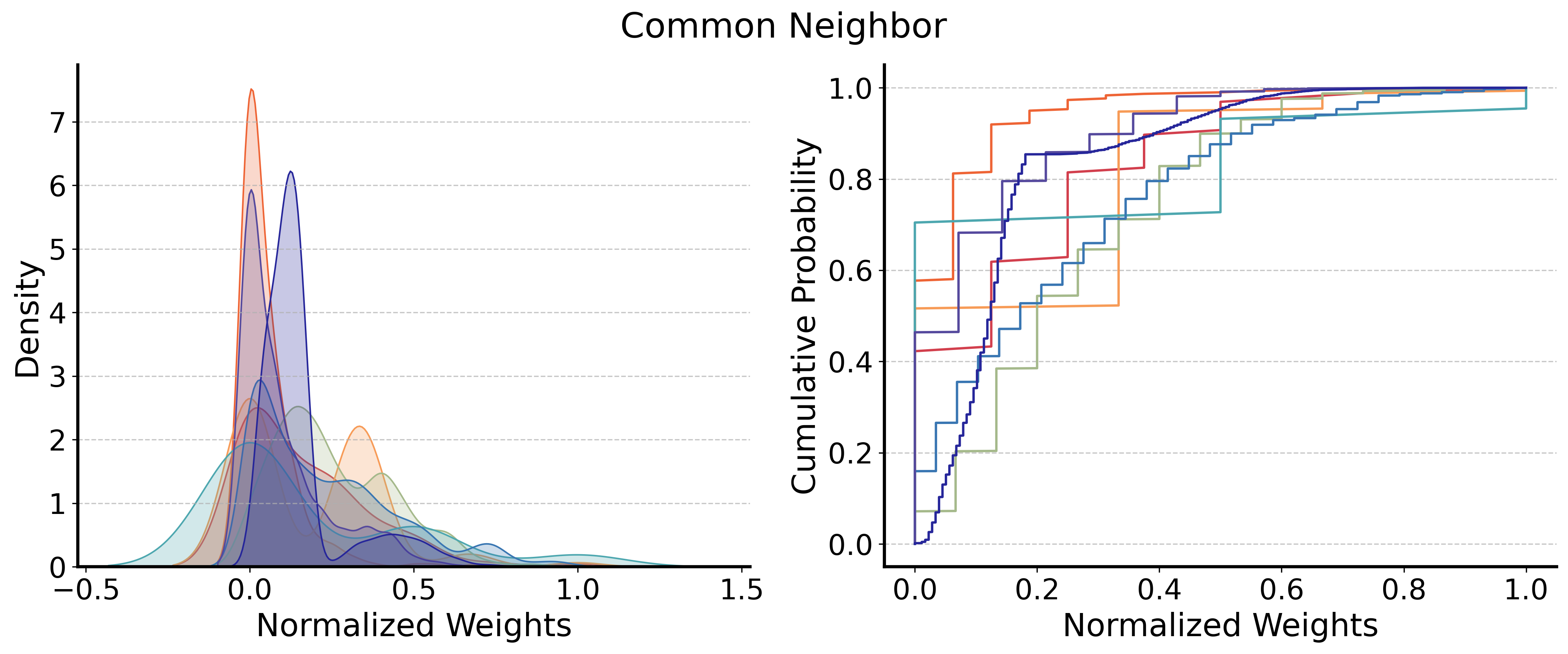}
		\caption*{} 
	\end{subfigure}\vspace{-18pt}
	\begin{subfigure}{0.46\textwidth}
		\centering
		\includegraphics[width=\linewidth]{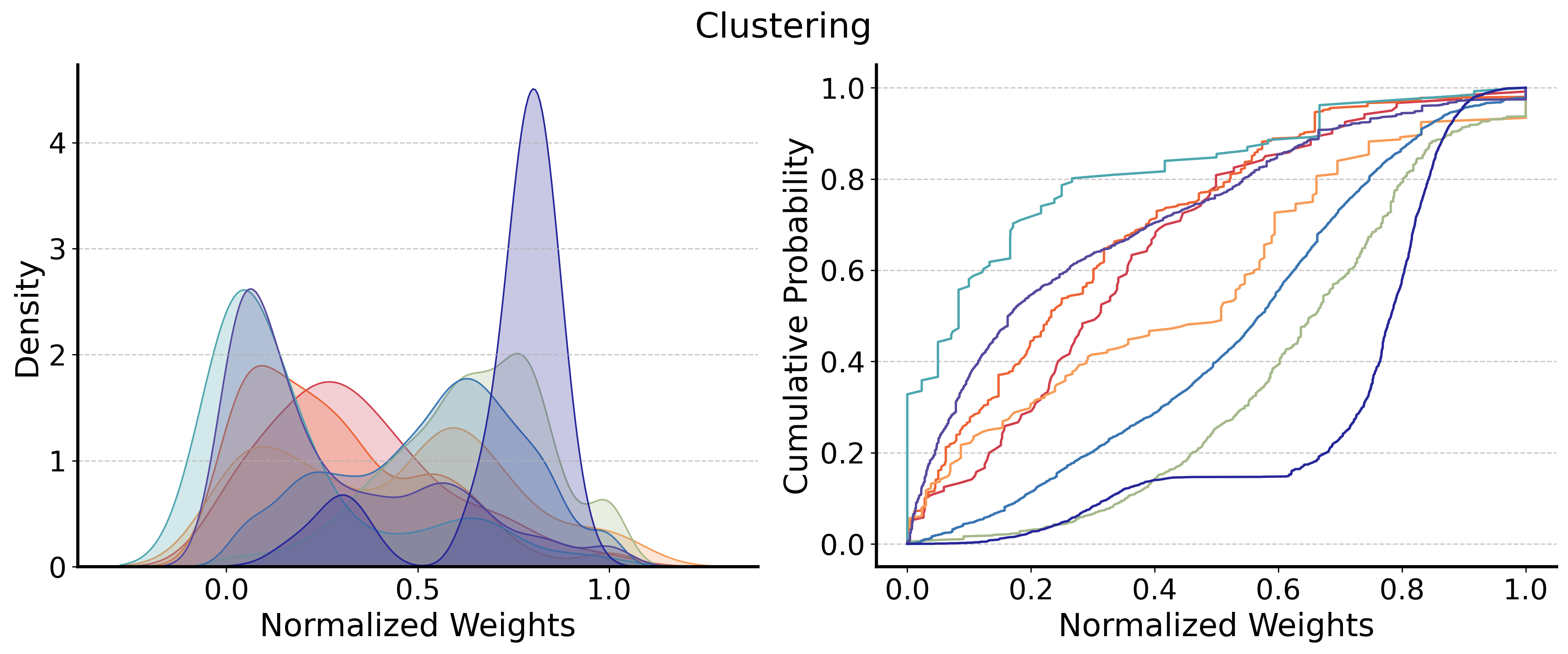}
		\caption*{} 
	\end{subfigure}
	\vspace{-30pt}
\end{figure}

\subsection{Limited Transferability of Single-Network Trained Models}
\begin{table*}[ht]
	\centering
	\caption{Prediction accuracy for each network when trained and tested on various networks. The diagonal values represent the accuracy of models trained and tested on the same network, while off-diagonal values show the transferability of models trained on one network to others. Underlined values indicate poor performance with accuracy below 0.6, highlighting the challenges in transferring models between networks with differing structural properties.}
	\label{tab:network_performance}
	\begin{tabular}{l|ccccc|ccccc|c}
		\hline
		Train / Test &  Airplane &  Fruit &  Thermody &  Weaver &  Worm &  Bacteria &  Coach &  Fungi &  Human &  WTW & \textbf{Average} \\
		\hline
		Airplane &    \textbf{0.9915} &         0.8034 &              \underline{0.5029} &       0.8149 &     0.7217 &        0.8619 &     0.9773 &     0.9011 &     0.7793 &   0.8498 &   0.8180 \\
		Fruit &        0.7249 &    \textbf{0.8589} &              \underline{0.5066} &       0.6293 &     \underline{0.5369} &        0.8777 &     0.8947 &     0.7374 &     0.7374 &   \underline{0.5827} &   0.7437 \\
		Thermody &        \underline{0.5524} &         \underline{0.4787} &    \textbf{0.7921} &       \underline{0.4827} &     \underline{0.5529} &        \underline{0.2632} &     0.5968 &     \underline{0.4298} &     \underline{0.4838} &   \underline{0.5469} &   0.5179 \\
		Weaver &        0.9427 &         0.7992 &              \underline{0.4909} &    \textbf{0.8704} &     0.6930 &        0.9017 &     0.9420 &     0.8718 &     0.7794 &   0.8135 &   0.8115 \\
		Worm &        0.6090 &         0.6824 &              \underline{0.4643} &       \underline{0.5509} &    \textbf{0.7529} &        0.7276 &     0.8207 &     0.6288 &     0.6712 &   \underline{0.4159} &   0.6326 \\
		\hline
		Average (Split) &        0.7641 &	0.7245&	0.5513&	0.6696&	0.6514&	0.7264&	0.8462&	0.7104&	0.6902&	0.6417 &   / \\
		\hline
	\end{tabular}
\end{table*}

Table~\ref{tab:network_performance} illustrates the performance of models trained on one network and tested on others. The results highlight two key findings:
First, models perform well on the same network they are trained on. Diagonal entries, such as Airplane (0.9915) and Fruit (0.8589), demonstrate the ability to accurately capture the relationship between network features and edge generation times within the same network.
Second, the models often fail when transferred to different networks. As seen in the underlined results, such as Thermody on Fruit (0.4787) and Fruit on Worm (0.5369), accuracies below 0.6 indicate poor transfer performance in binary classification tasks. This inconsistency reflects the complexity of real-world network mechanisms and the challenge of determining whether two networks are compatible for transfer.

In summary, models trained on a single network are unstable when applied to new networks. While some networks show decent transfer performance, others exhibit significant failures, highlighting the limitations of single-network training for generalizable predictions.

\subsection{Similarity Between Augmented and Original Networks}
This section evaluates the structural and dynamic similarity between augmented and the original networks. We employ various metrics, including degree distribution, eigenvector centrality, and clustering coefficient distribution, to assess similarity.
The analysis reveals that the degree distributions of the networks generated by TopoEvoDiff closely align with those of the original networks. This indicates a strong preservation of structural characteristics.
Furthermore, the eigenvector centrality measures show a robust correlation between nodes' importance in both networks, confirming the retention of critical structural roles.
Lastly, clustering coefficient distributions closely match between the generated and original networks, reflecting consistent local clustering tendencies.

Fig. \ref{fig:similarity} highlights the effectiveness of our approach, showing that the enhanced networks generated by TopoEvoDiff exhibit structures more similar to the original networks compared to those produced by the representative TIGGER method. The top graphs present results for the Airplane network, while the bottom graphs correspond to the Thermody network. This demonstrates the ability of our method to generate augmented networks with greater structural fidelity.

\subsection{Effect of Training Sample Size on Prediction Accuracy}
The results of predicting edge generation timestamps within a single network reveal an important finding: high prediction accuracy can be achieved using only a small subset of the network's edges. As shown in Fig. \ref{fig:train_ratio}, utilizing just 20\% of a network's edges for training yields strong performance, and increasing the number of training edges does not lead to significant improvements. This indicates that beyond a certain threshold, the additional training samples do not provide substantial value for improving accuracy within the same network.

This observation further supports the uniqueness of our augmentation strategy. The improvement in model performance on unseen networks is not merely a result of increasing the number of training samples. Instead, it demonstrates that the augmented samples generated through our method contribute meaningful and diverse information, enabling the model to generalize effectively across different networks. This contrasts with the plateauing effect observed when merely increasing training edges from the same network, highlighting the distinctive value of our augmentation approach.

\subsection{Benefits of Pairwise Relationship Prediction}\label{exp:benefit}
Pairwise relationship prediction addresses the challenges of network heterogeneity and facilitates cross-network generalization by normalizing relationships across networks. This approach is crucial for transferring learned models between networks with diverse structural and temporal characteristics, a task that direct prediction often fails to achieve.

Direct prediction struggles due to significant differences in edge timestamp distributions and feature characteristics across networks. For instance, as shown in Fig. \ref{fig:benefit}, edge timestamps in Thermody are evenly distributed, whereas other networks exhibit skewed distributions. Similarly, clustering coefficient distributions vary greatly, hindering generalization. 

Some network features are more transferable. For example, the common neighbor count exhibits similar distributions across networks, as demonstrated in Fig. \ref{fig:benefit}. This aligns with our experimental findings, where common neighbor count proves highly effective for learning the relationship between network features and edge generation timestamps. Furthermore, by shifting to pairwise learning of feature-edge relationships, normalization is achieved within the network itself, ensuring that the models learned through this method are inherently transferable across networks.

\section{Conclusion}

Our work focuses on predicting edge generation times from given network structures, which aids in understanding network evolution and forecasting future developmental trends.
Existing methods are capable of predicting the edge generation sequence for the remaining edges of a temporal network when part of the edges’ generation process is known. However, these methods often fail when applied to static networks. In real-world scenarios, a significant number of static networks, which lack timestamps, need to have their evolutionary processes reconstructed. To address this issue, we use the CPNN framework, which leverages the fusion of multiple temporal networks to train a model that can associate network structure with edge generation times across different networks. This approach achieves stable and improved accuracy in the prediction of evolution for unseen static networks. Furthermore, to further enhance the model’s stability and performance, we adopt the TopoEvoDiff generation model to produce additional temporal networks. By integrating real temporal networks with the generated ones for training, we achieve further improvements in the model’s prediction accuracy.

\bibliographystyle{IEEEtran}
\bibliography{tkde}

\vspace{-2cm}
\begin{IEEEbiography}[{\includegraphics[width=1in,height=1.25in,clip,keepaspectratio]{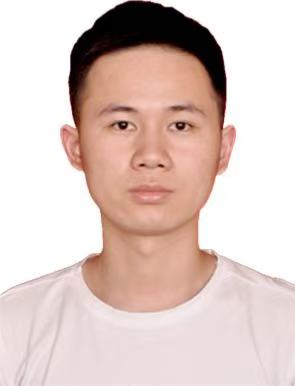}}]{En Xu}
received the BS and PhD degrees in Computer Science and Technology from Northwestern Polytechnical University, Xi'an, China, in 2018 and 2023, respectively. He was a postdoctoral research fellow at Hong Kong Baptist University from 2023 to 2024 and is currently a postdoctoral research fellow with the Department of Electronic Engineering, Tsinghua University, Beijing, China. His research interests include the predictability of time series and complex networks, with a current focus on AI for complex networks.
\end{IEEEbiography}

\vspace{-2cm}
\begin{IEEEbiography}[{\includegraphics[width=1in,height=1.25in,clip,keepaspectratio]{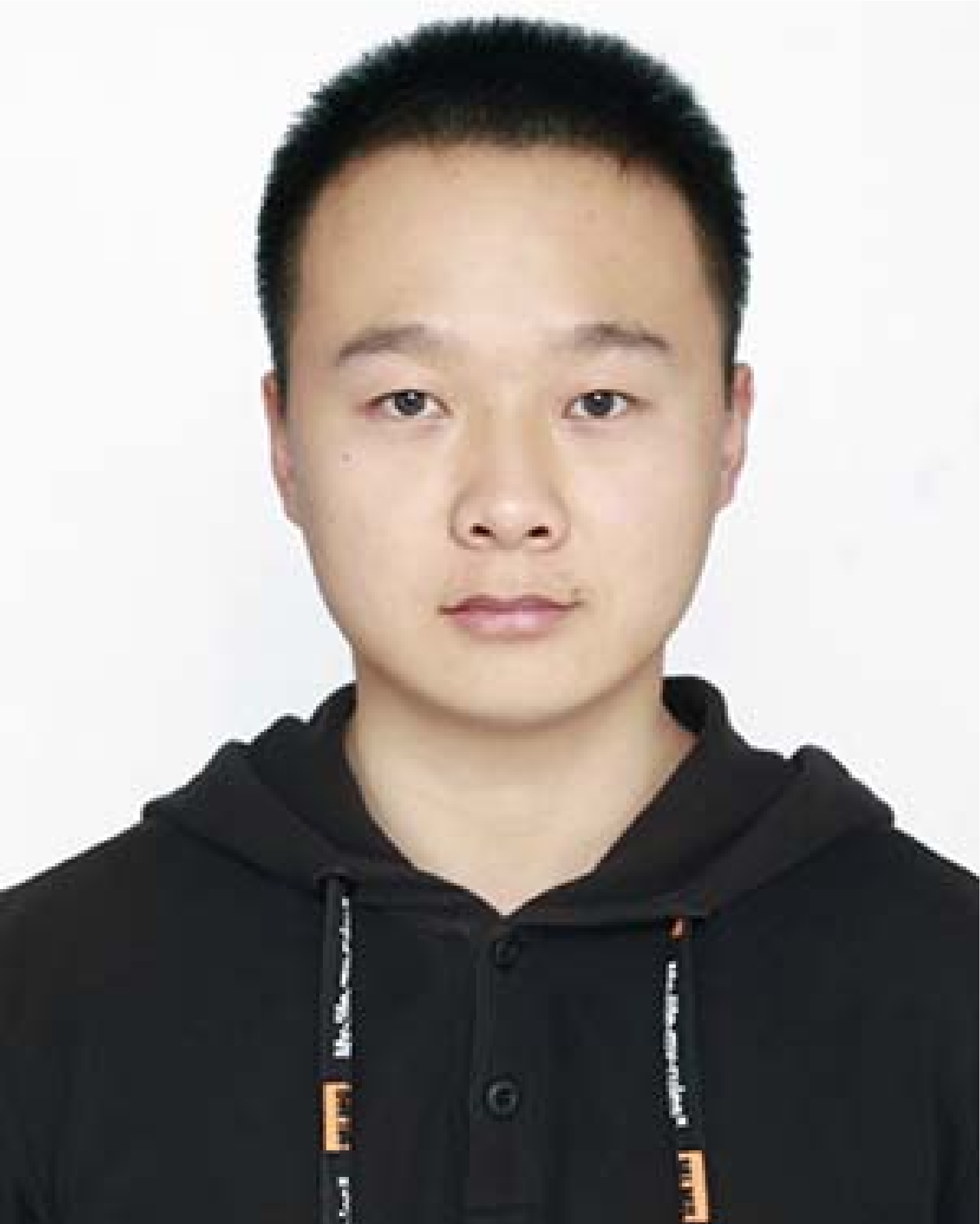}}]{Can Rong}
received the MEng degree in software engineering from Peking University in 2017. He is currently working toward the PhD degree with the Department of Electronic Engineering of Tsinghua University, advised by Prof. Yong Li. His research areas include urban computing and spatio-temporal data mining. He has recently been working on how to apply graph neural networks to model the interactive characteristics of regions in cities.
\end{IEEEbiography}

\vspace{-2cm}
\begin{IEEEbiography}[{\includegraphics[width=1in,height=1.25in,clip,keepaspectratio]{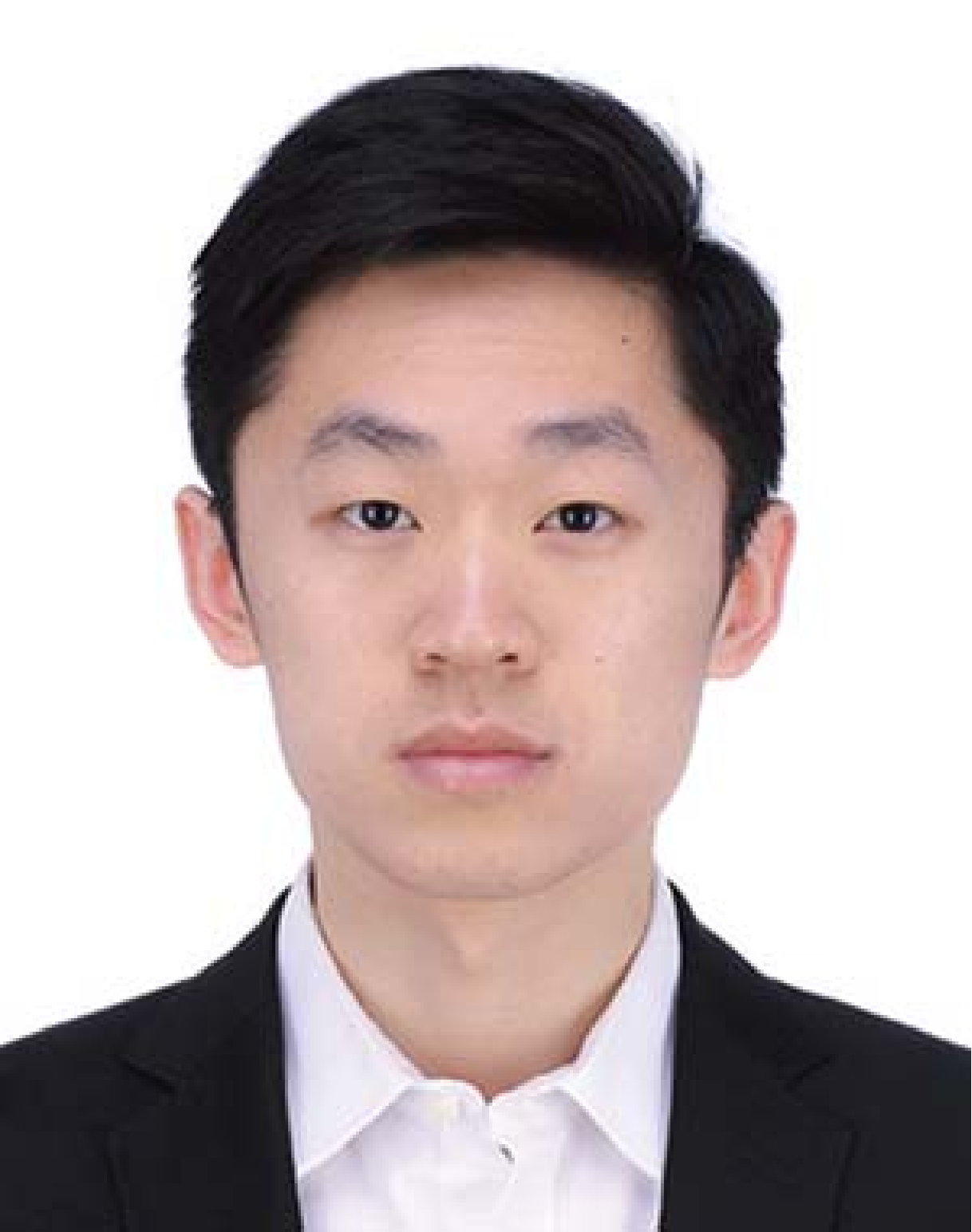}}]{Jingtao Ding}
received the BS degrees in electronic engineering and the PhD degree in electronic engineering from Tsinghua University, Beijing, China, in 2015 and 2020, respectively. He is currently a postdoctoral research fellow with the Department of Electronic Engineering, Tsinghua University. His research interests include mobile computing, spatiotemporal data mining and user behavior modeling. He has more than 60 publications in journals and conferences such as IEEE Transactions on Knowledge and Data Engineering, ACM Transactions on Information Systems, KDD, NeurIPS, WWW, ICLR, SIGIR, IJCAI, etc.
\end{IEEEbiography}

\vspace{-2cm}
\begin{IEEEbiography}[{\includegraphics[width=1in,height=1.25in,clip,keepaspectratio]{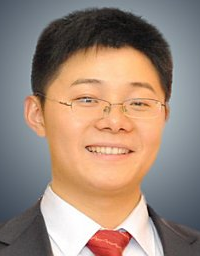}}]{Yong Li}
(Senior Member, IEEE) received the PhD degree in electronic engineering from Tsinghua University in 2012. He is currently a tenured professor with the Department of Electronic Engineering, Tsinghua University. He has authored or coauthored more than 100 papers on first-tier international conferences and journals, including KDD, WWW, UbiComp, SIGIR, AAAI, IEEE Transactions on Knowledge and Data Engineering, IEEE Transactions on Mobile Computing and his papers have total citations more than 28,000. His research interests include machine learning and big data mining, particularly, automatic machine learning and spatial-temporal data mining for urban computing, recommender systems, and knowledge graphs. He was the general chair, the TPC chair, an SPC or a TPC member for several international workshops and conferences. He is on the editorial board of two IEEE journals. Among them, 10 are ESI Highly Cited Papers in computer science, and five were the recipient of conference Best Paper (run-up) awards.
\end{IEEEbiography}

\appendix

\subsection{Equivalence Between Predicting Edge Generation Times and Pairwise Orderings}\label{apdx:equal}

In this section, we demonstrate the equivalence between predicting the exact generation times of network edges and predicting their pairwise order of generation. Consider a network with \(M\) edges generated sequentially. The normalized generation time sequence is denoted as \(\boldsymbol{\alpha} = (\alpha_1, \alpha_2, \ldots, \alpha_M)\), where \(\alpha_i = \frac{i}{M}\) represents the normalized position of edge \(i\) in the sequence, with larger values indicating later generation times. The goal is to reconstruct the sequence \(\boldsymbol{\alpha}\), either by directly estimating \(\alpha_i\) or by predicting the pairwise order of edges and deriving \(\boldsymbol{\alpha}\) from these comparisons.

For a pair of edges \(i\) and \(j\), the pairwise score \(u_{ij}\) is defined as:
\[
u_{ij} =
\begin{cases} 
	\frac{x}{M}, & \text{if } \alpha_i > \alpha_j, \\ 
	\frac{1-x}{M}, & \text{if } \alpha_i < \alpha_j,
\end{cases}
\]
where \(x\) denotes the probability of correctly predicting the order of a pair of edges. The expected value and variance of \(u_{ij}\) are given by:
\[
\mathbb{E}(u_{ij}) = 
\begin{cases} 
	\frac{x}{M}, & \text{if } \alpha_i > \alpha_j, \\ 
	\frac{1-x}{M}, & \text{if } \alpha_i < \alpha_j, 
\end{cases}
\]
\[
\text{Var}(u_{ij}) = \frac{x(1-x)}{M^2}.
\]

Using the pairwise scores, the total score \(u_i\) for edge \(i\) is calculated as:
\[
u_i = \sum_{j=1, j \neq i}^M u_{ij}.
\]

The expected score of edge \(i\) is determined by summing its comparisons with all other edges. Specifically, comparisons with the first \(i-1\) edges contribute \((i-1) \frac{x}{M}\), while comparisons with the remaining \(M-i\) edges contribute \((M-i) \frac{1-x}{M}\). The total expected score is therefore:
\[
\mathbb{E}(u_i) = (i-1) \frac{x}{M} + (M-i) \frac{1-x}{M} = \frac{2x-1}{M} i + \frac{1-x}{M}.
\]

This expression shows that \(\mathbb{E}(u_i)\) is a linear function of the edge index \(i\). The scores are evenly distributed over the interval \([1-x, x]\), allowing \(u_i\) to serve as a mean-field approximation for ranking edges. The normalized position \(\hat{\alpha}_i\) of edge \(i\) in the reconstructed sequence can be calculated as:
\[
\hat{\alpha}_i = \frac{u_i - (1-x)}{2x-1}.
\]

The variance of \(\hat{\alpha}_i\) can be derived as:
\[
\text{Var}(\hat{\alpha}_i) = \frac{\text{Var}(u_{ij})}{(2x-1)^2} = \frac{x(1-x)}{(2x-1)^2} \frac{1}{M},
\]
and the standard deviation is:
\[
\text{Std}(\hat{\alpha}_i) = \sqrt{\frac{x(1-x)}{(2x-1)^2}} \frac{1}{\sqrt{M}}.
\]

The expected value of \(\hat{\alpha}_i\) matches the true value \(\alpha_i\), demonstrating that the reconstructed sequence is unbiased:
\[
\mathbb{E}(\hat{\alpha}_i) = \alpha_i.
\]

The overall theoretical error in reconstructing the sequence is equivalent to the standard deviation of the position estimates:
\[
\mathcal{E}_\text{theory} = \sqrt{\frac{x(1-x)}{(2x-1)^2}} \frac{1}{\sqrt{M}}.
\]

This theoretical error decreases as the pairwise accuracy \(x\) approaches 1 or as the number of edges \(M\) increases. However, the error diverges as \(x\) approaches 0.5, emphasizing the importance of achieving reliable pairwise predictions. This equivalence highlights that predicting pairwise order is sufficient to reconstruct the temporal sequence of edges, offering a simpler yet equally effective alternative to directly predicting generation times.

\subsection{Reconstructing Edge Generation Order}\label{apdx:borda}

The following algorithm reconstructs the global edge generation order from pairwise comparison results. The pairwise comparison matrix \(P\) contains the probabilities \(P_{ij}\), where \(P_{ij}\) indicates the likelihood that edge \(i\) was generated before edge \(j\). The algorithm computes a score for each edge based on these pairwise probabilities and then sorts the edges by their scores to derive the global order.
\begin{algorithm}
	\renewcommand{\algorithmicrequire}{\textbf{Input:}}
	\renewcommand{\algorithmicensure}{\textbf{Output:}}
	\caption{Reconstructing Edge Generation Order using Pairwise Comparisons}
	\label{alg:borda_reconstruction}
	\begin{algorithmic}[1]
		\REQUIRE Pairwise probability matrix \(P\) of size \(E \times E\), where \(P[i][j]\) represents the probability that edge \(i\) is generated before edge \(j\).
		\ENSURE Reconstructed edge generation order \(\hat{\alpha}\).
		\STATE Initialize a score vector \(S\) of size \(E\), with all elements set to 0.
		\FOR{each edge \(i \in \{1, \dots, E\}\)} 
		\FOR{each edge \(j \in \{1, \dots, E\}, j \neq i\)}
		\STATE Update \(S[i] = S[i] + P[i][j]\).
		\ENDFOR
		\ENDFOR
		\STATE Sort edges by \(S\) in ascending order to obtain \(\hat{\alpha}\).
		\RETURN \(\hat{\alpha}\).
	\end{algorithmic}  
\end{algorithm}

\subsection{Validation of Augmentation Strategy in Separate Training}\label{apdx:enhance}

This section evaluates the effectiveness of combining a single original network with an augmented network in predicting edge generation times for other networks. The experiments were conducted on three types of generated datasets, and the detailed descriptions of these datasets are as follows:
\begin{itemize}
	\item Barabási–Albert (BA) model: The BA model generates networks through \textit{preferential attachment}. Starting with a small fully connected graph, each new node added to the network connects to an existing node with probability $p \propto k$, where $k$ is the degree (number of connections) of the existing node. This model reflects the \textit{power-law degree distribution} found in many natural networks, where a few nodes dominate with many connections.
	
	\item Popularity-similarity-optimization (PSO) model: The PSO model introduces a similarity space for nodes, where the connection probability between new and existing nodes depends on both the degree of the target node and their similarity. This model is well-suited for simulating social networks, where users (nodes) connect based on both popularity and shared interests.
	
	\item Fitness model: In the Fitness model, each node is assigned a \textit{fitness value} $\eta_i$, representing its inherent attractiveness or importance. The probability of a new node connecting to an existing node is proportional to both the degree and fitness of the target node, $p_{ij} \propto k_j \eta_j$. This model captures networks with strong competitive dynamics, such as academic citation networks or business networks.
\end{itemize}

The network types $\textup{BA}_{100}$, $\textup{Fitness}_{100}$, $\textup{PSO}_{100}$, and $\textup{PSO}_{200}$ represent different network models, where the subscript indicates the number of nodes in the network. TrainingNet refers to the network used for standalone training, based on which enhanced samples are generated. Subsequently, the combination of this network and one enhanced sample is used to predict the target network (TargetNet). For the methods, NatureC is the baseline method presented in Nature Communications 2024 \cite{wang2024reconstructing}. TIGGER combines NatureC with enhanced networks generated by TIGGER, while TopoEvoDiff integrates NatureC with enhanced networks generated by TopoEvoDiff. The improvements achieved through augmentation are summarized in Table \ref{tab:enhance_single}, which demonstrates the effectiveness of our augmentation strategy in enhancing prediction performance.
\begin{table}[htbp]
	\centering
	\caption{Effectiveness of enhancement methods on synthetic datasets.}
	\label{tab:enhance_single}
	\resizebox{0.49\textwidth}{!}{
		\begin{tabular}{llccc}
			\toprule
			TrainingNet & TargetNet & NatureC & TIGGER & TopoEvoDiff  \\
			\midrule
			$\textup{BA}_{100}$     & $\textup{Fitness}_{100}$     & 0.5189  & 0.5368 (+3.4\%)  & 0.5478 (+5.6\%) \\
			$\textup{Fitness}_{100}$ & $\textup{BA}_{100}$        & 0.5148  & 0.5254 (+2.1\%)  & 0.5448 (+5.8\%)   \\
			$\textup{PSO}_{100}$    & $\textup{PSO}_{200}$         & 0.6103  & 0.6267 (+2.7\%)  & 0.6447 (+5.6\%)  \\
			$\textup{PSO}_{200}$    & $\textup{PSO}_{100}$          & 0.5562  & 0.5778 (+3.9\%)  & 0.6281 (+12.9\%) \\
			\bottomrule
		\end{tabular}
	}
\end{table}

\end{document}